\documentclass[twocolumn]{svjour3}
\smartqed  

\usepackage{graphicx}
\usepackage{amsmath}
\usepackage{amssymb}
\usepackage{tabularx}
\usepackage{multirow}
\usepackage{algorithm}
\usepackage{algorithmic}
\usepackage{stackengine}
\usepackage{xcolor}
\usepackage{adjustbox}
\usepackage{colortbl}
\usepackage{ragged2e}
\usepackage{rotating,soul}
\usepackage[caption=false,font=footnotesize]{subfig}
\usepackage[left=1.8cm,right=1.8cm,top=2.0cm,bottom=2.7cm]{geometry}
\usepackage[colorlinks,linkcolor=green,citecolor=blue,urlcolor=blue]{hyperref}

\usepackage{setspace}
\usepackage[numbers,sort]{natbib}
\bibpunct[,]{(}{)}{;}{a}{}{,}
\renewcommand{\cite}[1]{\textcolor{blue}{\citep{#1}}}

\usepackage{pifont}
\newcommand{\xmark}{\ding{55}}%

\graphicspath{{./images/}}
\def\eg{\emph{e.g.}}
\def\ie{\emph{i.e.}}

\newcommand{\figref}[1]{Fig.~\ref{#1}}
\newcommand{\tabref}[1]{Tab.~\ref{#1}}
\newcommand{\equref}[1]{Eqn.~\ref{#1}}
\newcommand{\secref}[1]{Sec.~\ref{#1}}

\newcommand{\addImg}[2][0.495]{\includegraphics[width=#1\textwidth]{#2}}

\journalname{International Journal of Computer Vision}

\begin{document}

\title{Unsupervised Scale-consistent Depth Learning from Video}

\author{Jia-Wang Bian$^{1,2}$ \and
 		Huangying Zhan$^{1,2}$ \and
    Naiyan Wang$^3$ \and 
    Zhichao Li$^3$ \and
    Le Zhang$^4$ \and
    Chunhua Shen$^{1,2}$ \and
    Ming-Ming Cheng$^5$ \and
    Ian Reid$^{1,2}$
}


\authorrunning{International Journal of Computer Vision (2021)}

\institute{
$^1$ The University of Adelaide, Australia \\
$^2$ Australian Centre for Robotic Vision, Australia \\
$^3$ TuSimple, China \\
$^4$ Agency for Science, Technology and Research, Singapore \\
$^5$ TKLNDST, CS, Nankai University, China \\
}

\date{Received: 7 September 2020 / Accepted: 24 May 2021}

\maketitle

\begin{abstract}
We propose a monocular depth estimator \emph{SC-Depth},
which requires only unlabelled videos for training and enables the scale-consistent prediction at inference time.
Our contributions include: (i) we propose a geometry consistency loss, 
which penalizes the inconsistency of predicted depths between adjacent views;
(ii) we propose a self-discovered mask to automatically localize moving objects that violate the underlying static scene assumption and cause noisy signals during training;
(iii) we demonstrate the efficacy of each component with a detailed ablation study and show high-quality depth estimation results in both KITTI and NYUv2 datasets.
Moreover, thanks to the capability of scale-consistent prediction,
we show that our monocular-trained deep networks are readily integrated into ORB-SLAM2 system for more robust and accurate tracking.
The proposed hybrid Pseudo-RGBD SLAM shows compelling results in KITTI,
and it generalizes well to the KAIST dataset without additional training.
Finally, we provide several demos for qualitative evaluation.
The source code 
 is 
 released on \href{https://github.com/JiawangBian/SC-Depth-Release}{GitHub}.

\end{abstract}

\section{Introduction}

\begin{figure}[t]
  \centering
  \addImg[0.46]{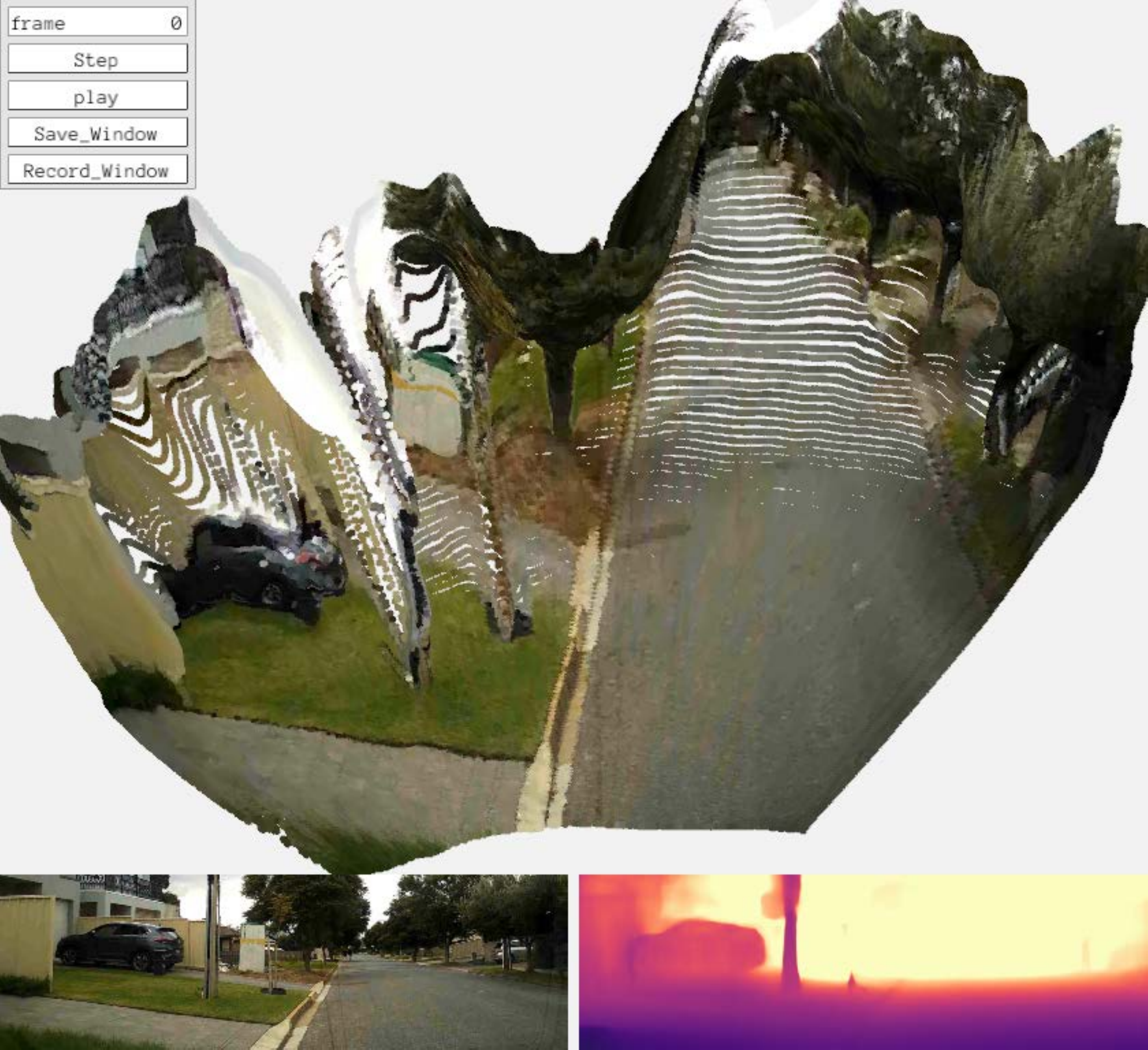} \\ 
  (a) Predicted depth and textured point cloud \\ 
  \addImg[0.46]{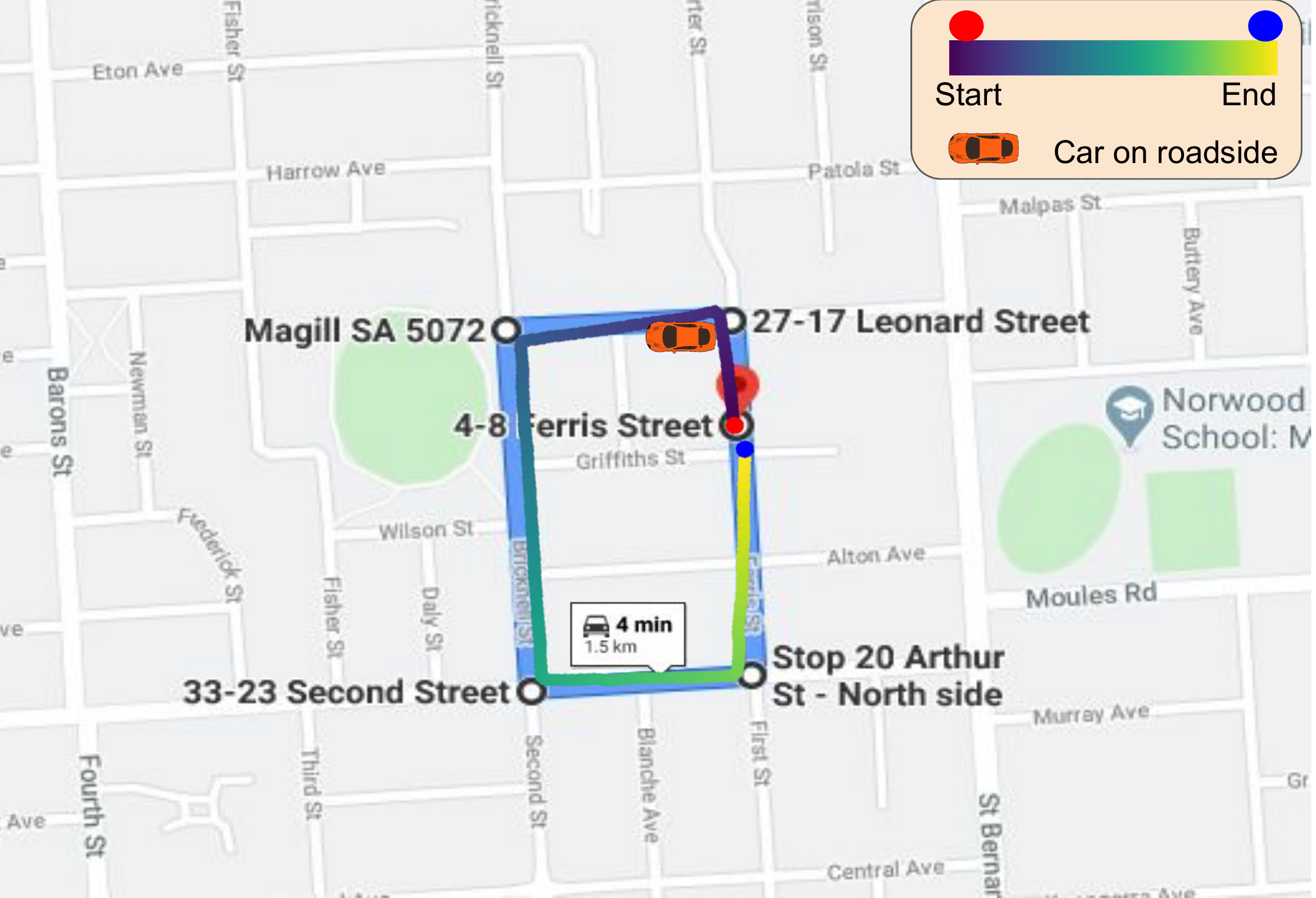} \\
  (b) Qualitative evaluation of estimated trajectory 
  \caption{Generalization on our self-captured video. The data is collected in Adelaide, an Australia city. Our depth and pose networks are trained on KITTI without additional finetuning. The scene is so challenging that ORB-SLAM2~\cite{murORB2} failed to initialize or quickly lost tracking after initialization, while our Pseudo-RGBD SLAM system can provide an accurate trajectory, which is consistent with the Google Map. See more details in \secref{sec:demo}.
  }\label{fig:demo-adl}
\end{figure}

The CNN-based monocular depth estimation~\cite{eigen2014depth} has shown significant promise for many Computer Vision tasks.
The supervised methods~\cite{fu2018deep, Yin2019enforcing} achieve high performance,
while they require expensive range sensors to capture the ground-truth data for training.
To this end, recent work explores unsupervised learning for monocular depth estimation,
which either uses the calibrated stereo pairs~\cite{garg2016unsupervised,godard2017unsupervised} or unlabelled videos~\cite{zhou2017unsupervised, yin2018geonet} for training.
In these frameworks, the color consistency between multiple views serves as the main supervision signal.
Since they do not require ground truth labels, and particularly the recent method ~\cite{gordon2019depth} showing unsupervised depth estimation can work with the unknown camera intrinsics,
these methods attract a lot of interest in the Computer Vision community.
In this paper, we are interested in the video-based unsupervised learning framework because it has a minimum requirement for training data. 

Compared with stereo-based learning,
video-based learning~\cite{zhou2017unsupervised} is often more challenging due to the unknown camera motion.
More importantly, due to \emph{scale ambiguity}, a natural issue in monocular vision, 
the predicted depth by the latter has an unknown scaling to the real world.
This is the so-called relative depth, as opposed to the metric depth in the previous setting.
The relative depth is also widely used,
\eg, O\-R\-B-S\-L\-A\-M \cite{mur2015orb} and COLMAP~\cite{schonberger2016structure} both generate results up to an unknown scale.
However, one critical issue that we identify in video-based learning is that methods may generate \emph{scale-inconsistent} predictions over different frames
since they suffer from a per-frame scale ambiguity.
This does not impact the single-image based tasks,
while it is critical for video-based applications,
\eg, inconsistent depths cannot be used for camera tracking in the Visual SLAM system---See \figref{fig:keypoints}.

In this paper, we propose an improved unsupervised learning framework for higher depth accuracy and consistency.
First, we propose a geometry consistency loss ($L_{G}$) to encourage networks to predict scale-consistent depths.
It explicitly penalizes the pixel-wise inconsistency of predicted depths between adjacent frames during training.
It enables more effective learning and allows for more consistent predictions at inference time---See \tabref{tab:consistency}.
Second, we propose a self-discovered mask ($M_s$) for handling moving objects during training,
which violates the underlying static scene assumption.
It improves the performance significantly (See \tabref{tab:ablation}) and does not require additional overhead since the proposed mask is simply derived from $L_{G}$.

To show the benefits from scale-consistent depth prediction and demonstrate our contribution for downstream tasks,
we integrate our trained networks into the ORB-SLAM2~\cite{murORB2} system for more accurate and robust tracking.
The proposed hybrid Pseudo-RGBD SLAM system has distinct advantages over traditional monocular systems,
including
\textbf{a)} it starts tracking at any frame without latency;
\textbf{b)} it enables more robust and accurate tracking with the help of predicted depths;
and \textbf{c)} it allows for dense 3D reconstruction---See \figref{fig:demo-kitti}.
We report comprehensive quantitative results and provide several demos in \secref{sec:demo} for qualitative evaluation.
An example is shown in \figref{fig:demo-adl},
where we visualize the depth, point cloud, and camera trajectory generated by our method on a real-world driving video.

Our preliminary version was presented in NeurIPS 2019~\cite{bian2019depth},
where we propose an unsupervised learning framework for scale-consistent depth and pose estimation.
In this paper, we \textbf{i)} add more technical details of our proposed method;
\textbf{ii)} make a more clear explanation of our contribution and distinguish our method from existing methods;
\textbf{iii)} improve our learning framework by changing network architectures and integrating effective components from related work;
\textbf{iv)} conduct a more comprehensive evaluation and show the potential of our method to Visual SLAM. 

\section{Related work}

\paragraph{Single-view depth estimation.}
The depth estimation problem was mainly solved by using traditional geometry based methods~\cite{Geiger2011IV, schoenberger2016mvs} before deep learning based methods emerged.
They rely on correspondences search~\cite{lowe2004distinctive, Bian2019gms}, model fitting~\cite{zhang1998determining, bian2019bench}, and multi-view triangulation~\cite{hartley2003multiple}.
Therefore, at least two different views of the scene are required as input for computing the depth.
In contrast, recent deep learning based methods leverage the expressive power of convolutional neural networks,
and they are able to regress the depth from a single image only.
According to the training data, we can categorize learning-based methods into four classes:
First, \cite{eigen2014depth} use the sensor captured depths  (\eg, LiDAR or RGB-D devices) as the ground truth for training.
The following work~\cite{liu2016learning,fu2018deep,garg2019learning,Yin2019enforcing,huynh2020guiding} proposes more advanced network architectures or learning objectives to improve the performance.
These methods achieve high performance, while it is expensive to capture ground-truth data in many real-world scenes.
Second, \cite{xian2018monocular,li2018megadepth,li2019learning,wang2019web,chen2019learning,yin2020learning} collect stereo images or videos from the web and use off-the-shelf tools (\eg, stereo matching~\cite{hirschmuller2005accurate} or multi-view stereo~\cite{schoenberger2016mvs}) to compute dense ground-truth depths.
Besides, \cite{ranftl2020towards} export perfect depths from the synthetic 3D movies~\cite{Butler_ECCV_2012}.
Although these methods can obtain cheap ground-truth data,
there often exists a domain gap between the collected data and the desired scenes.
More importantly, the learned scale information is hard to generalize across different scenes so that they often predict the relative depth.
This prevents them from predicting consistent depths on a video.
Third, \cite{garg2016unsupervised} use the calibrated stereo images for training models,
where they warp images using the predicted depth with the known camera baseline and use the photometric loss to penalize the warping error.
Then \cite{godard2017unsupervised} exploit the left-right consistency in image pairs,
and \cite{zhan2018unsupervised} exploit the temporary consistency in videos.
\cite{pilzer2018unsupervised} leverage adversarial learning,
and \cite{Poggi_CVPR_2020} study the uncertainty of predicted depths.
These methods can predict the metric depth,
while it requires well-calibrated stereo cameras to collect training data.
Fourth, \cite{zhou2017unsupervised} train models from unlabelled videos,
where they jointly train the depth and pose networks using adjacent frames with photometric loss and differentiable warping~\cite{jaderberg2015stn}.
Due to the simplicity and generality,,
it attracts a lot of researchers' interests and inspires a series of works,
including \cite{mahjourian2018unsupervised, Wang2018CVPR, yin2018geonet, zou2018df, ranjan2019cc, monodepth2, chen2019self, gordon2019depth, zhao2020towards, Zhou_2019_ICCV,klingner2020self,packnet,packnet-semguided}.
Our method falls into this category,
and we target improving the depth accuracy and consistency for advancing downstream video-based tasks.

\paragraph{Scale consistency.}
It is an important problem in Visual SLAM~\cite{mur2015orb},
but to the best of our knowledge, 
we are the first ones to discuss the scale inconsistency behind unsupervised video-based depth learning.
Nevertheless, we find that our proposed geometry consistency loss is technically similar to two previous methods.
First, \cite{mahjourian2018unsupervised} propose a 3D ICP loss to penalize the misalignment of predicted depths,
where they approximate gradients for depth and pose networks independently because the ICP is not differentiable.
This ignores second-order effects between depth and pose networks,
and hence it limits the performance.
By contrast, our geometry consistency loss is naturally differentiable and results in better performance.
Second, \cite{zou2018df} propose a depth consistency loss,
which enforces corresponding points in two images to have identical depth predictions.
This is physically incorrect because the scene depth is view-dependent,
\ie, it should be different in different views.
We instead synthesize the depth for the second view using the predicted depth in the first view via rigid transformation,
and we penalize the difference between predicted depths and synthesized depths in the second view.
Not only does our approach improve the depth accuracy, 
but also it enables scale-consistent depth prediction for advancing video-based applications such Visual SLAM~\cite{murORB2}.
After the publication of our conference paper,
we notice that more recent works pay attention to consistent depth prediction,
including~\cite{luo2020consistent,tiwari2020pseudo,zhao2020towards,zou2020learning}.

\paragraph{Moving objects.}
As the moving objects violate the underlying static world assumption for learning depths,
related work often detects dynamic regions and masks them out when computing the photometric loss.
\cite{zhou2017unsupervised} predict a mask from a pair of images by using the neural network.
However, due to lacking effective supervision signals, the performance is limited.
\cite{vijayanarasimhan2017sfm} learn a moving object mask from synthetic data~\cite{menze2015object},
which is often hard to generalize to real-world scenes.
\cite{yin2018geonet, zou2018df, ranjan2019cc, chen2019self} additionally train an optical flow network and compare the optical flow with depth-based mapping for detecting moving objects.
This is effective, but training an optical flow network is time-consuming due to the complex correlation computation.
\cite{casser2019struct2depth,gordon2019depth,packnet-semguided,huynh2020guiding,casser2019unsupervised} leverage the semantic information for localizing dynamic objects.
They either require the pretrained semantic segmentation network or need the manually labelled class labels for multi-task training.
\cite{monodepth2} mask out the moving objects that have the same velocity as the camera,
while it cannot handle other object motions.
Compared with previous methods, our method does not require semantic inputs and does not require training additional networks.
Our proposed mask is analytically derived from the geometry consistency loss,
and it is able to handle arbitrary object motions and occlusions.
After ours, \cite{li2020unsupervised} propose to learn the dense 3D translation field of objects relative to the scene by using the neural network,
which is also efficient and effective.

\paragraph{Depth estimation for Visual SLAM.}
Traditional methods use either feature matching~\cite{Geiger2011IV,klein2007parallel} or direct image alignment~\cite{forster2014svo, engel2017direct} for camera tracking and mapping. 
Recently, \cite{yin2017scale} use a supervised depth estimation model to help recover the absolute scale for monocular methods.
CNN-SLAM~\cite{tateno2017cnn} uses the depth estimation network within a monocular SLAM system for dense reconstruction.
CodeSLAM~\cite{bloesch2018codeslam} jointly optimizes the depth and pose via a learned latent code.
Although promising results are reported,
these methods rely on supervised training,
which is not always available in real-world scenarios.
UndeepVO~\cite{li2018undeepvo} and \cite{zhan2018unsupervised} train depth and pose networks on the calibrated stereo videos using the photometric loss,
and they show that the learned pose network can inference on monocular videos like a visual odometer.
CNN-SVO~\cite{loo2019cnn} combines the stereo-learned depth network and SVO~\cite{forster2014svo} for more accurate trajectory estimation. 
DVSO~\cite{yang2018deep} and D3VO~\cite{yang2020d3vo} also train depth models on stereo videos,
and they further conduct geometric optimization.
Note that all the aforementioned methods do not suffer from the scale ambiguity issue, as opposed to ours,
because they can recover the metric depth.
In this paper, we show that the monocular-trained model can predict the scale-consistent results,
and it can be used for visual odometry.
After ours, \cite{zou2020learning} propose to model the long-term dependency by using a two-layer convolutional LSTM module,
which improves the pose prediction accuracy significantly.
However, the pure learning-based methods are easy to overfit,
and we believe that combing deep learning and geometry-based methods is a more promising direction.
As a result, our hybrid system generalizes well to the previously unseen dataset and to our self-captured videos.

\begin{figure*}[ht]
  \centering
  \addImg[1]{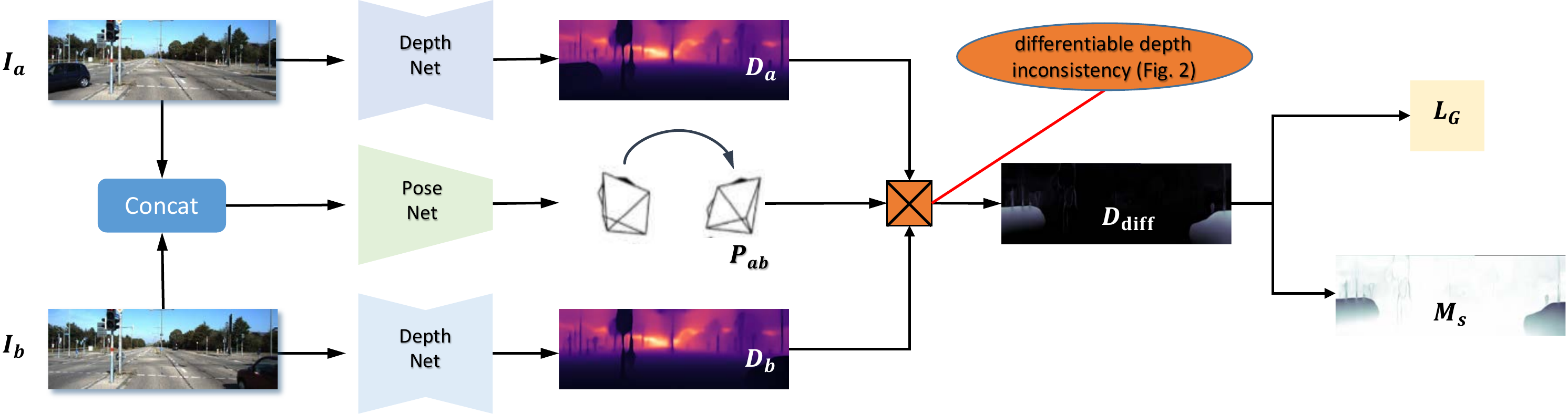}
  \caption{Illustration of the proposed geometry consistency loss and self-discover mask.
    Given two consecutive frames ($I_a$, $I_b$), 
    we first estimate their depth maps ($D_a$, $D_b$) and relative pose ($P_{ab}$) using the network.
    Then we compute the $D_{\text{diff}}$ (\equref{eqn-depthdiff}), \ie, pixel-wise depth inconsistency between $D_a$ and $D_b$.
    Finally, we derive our geometric consistency loss $L_{G}$ (\equref{eqn-gc}) and self-discovered mask $M_s$ (\equref{eqn-mask})
    from $D_{\textit{diff}}$ to regularize the network training and hanlding dynamics and occlusions (\figref{fig:show}).
    For clarity, the photometric loss and smoothness loss are not shown in 
    this figure.}
    \label{fig:sc-framework}
\end{figure*}

\begin{figure}[ht]
  \centering
  \includegraphics[width=\linewidth]{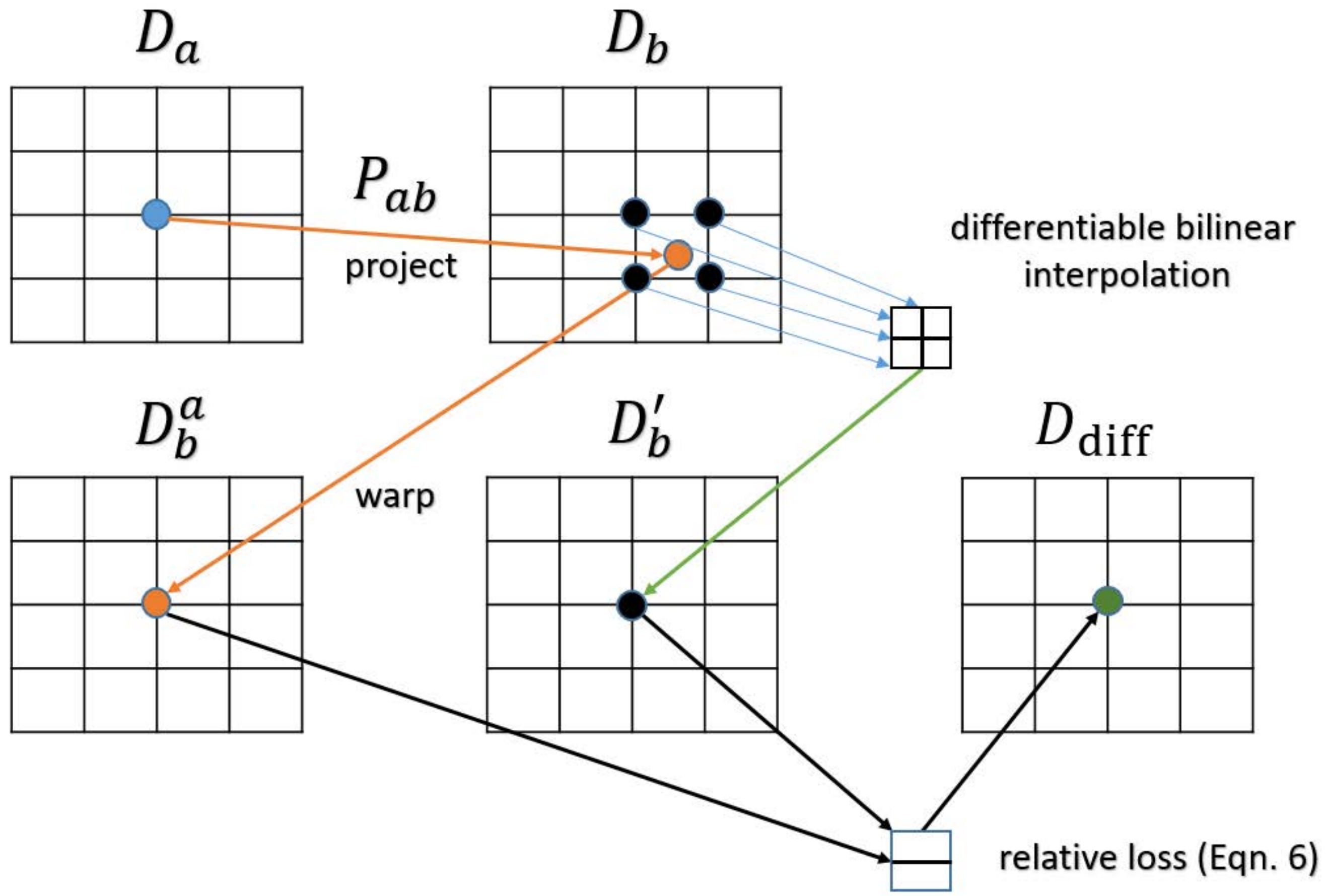}
  \caption{Differentiable depth inconsistency computation. 
  This operation takes two depth maps ($D_a$, $D_b$) and their relative pose ($P_{ab}$) as input and outputs the pixel-wise inconsistency.
  Firstly, we project $D_a$ to 3D space and then to the image plane of $D_b$ using $P_{ab}$,
  obtaining the $D^a_b$ that stands for the synthesized $D_b$.
  Then, we hope to compute the difference between $D^a_b$ and $D_b$.
  However, it is not practical because the projection does not religiously lie in the grid of $D_b$.
  Therefore, we obtain the $D'_b$ by using the \emph{differentiable bilinear interpolation}~\cite{jaderberg2015stn}.
  Finally, we compare $D^a_b$ with $D'_b$ to obtain the depth inconsistency ($D_{\text{diff}}$).
  Here, we use the relative loss (\equref{eqn-depthdiff}), although other loss functions such as L1 and L2 are also applicable.
  }\label{fig:gc-loss}
\end{figure}

\section{SC-Depth}\label{sec-learning}

\subsection{Framework Overview}\label{sec-overview}

Our goal is to train depth and pose CNNs from unlabeled videos. 
Given two adjacent frames ($I_a$, $I_b$) randomly sampled from a video,
their depth maps ($D_a$, $D_b$) and relative 6-DoF camera pose $P_{ab}$ 
are first estimated by the depth and pose CNNs, respectively.
With the predicted depth and pose, 
we can synthesize the reference image $I_a$ using the source image $I_b$ by differentiable warping~\cite{jaderberg2015stn},
which generates $I_a'$.
Then the network is supervised by the photometric loss between 
the real $I_a$ and the synthesized $I_a'$.
To explicitly constrain the depth CNN to predict scale-consistent results on adjacent frames,
we propose a geometry consistency loss $L_{G}$.
To handle invalid cases such as static frames and dynamic objects, we introduce two masks. 
First, a self-discovered mask $M_s$ (\equref{eqn-mask}) is introduced to reason the dynamics and occlusions by checking the depth consistency. 
\figref{fig:sc-framework} illustrates the proposed loss and mask.
Second, we use the auto-mask ($M_a$)~\cite{monodepth2} to remove stationary points on image pairs where the camera is not moving.

Our objective function is formulated as follows:
\begin{equation}\label{eqn-totalloss}
  L = \alpha L_{P}^M + \beta L_{S} + \gamma L_{G},
\end{equation}
where $L_{P}^M$ stands for the photometric loss $L_{P}$ weighted by the proposed $M_s$. 
$L_{S}$ stands for the smoothness loss, 
and $L_{G}$ is the geometric consistency loss.
$[\alpha, \beta, \gamma]$ are the loss weighting terms.
The loss is averaged over valid points, which are determined by $M_a$.
In the following sections, we first introduce the photometric loss and smoothness loss in \secref{sec-photo-smooth}, 
then we describe the proposed geometric consistency loss $L_{G}$ in \secref{sec-gc} and the self-discovered mask $M_s$ in \secref{sec-mask},
and finally, we elaborate the auto-mask $M_a$ in \secref{sec-auto-mask}.

\begin{figure*}[t]
  \centering
  \addImg{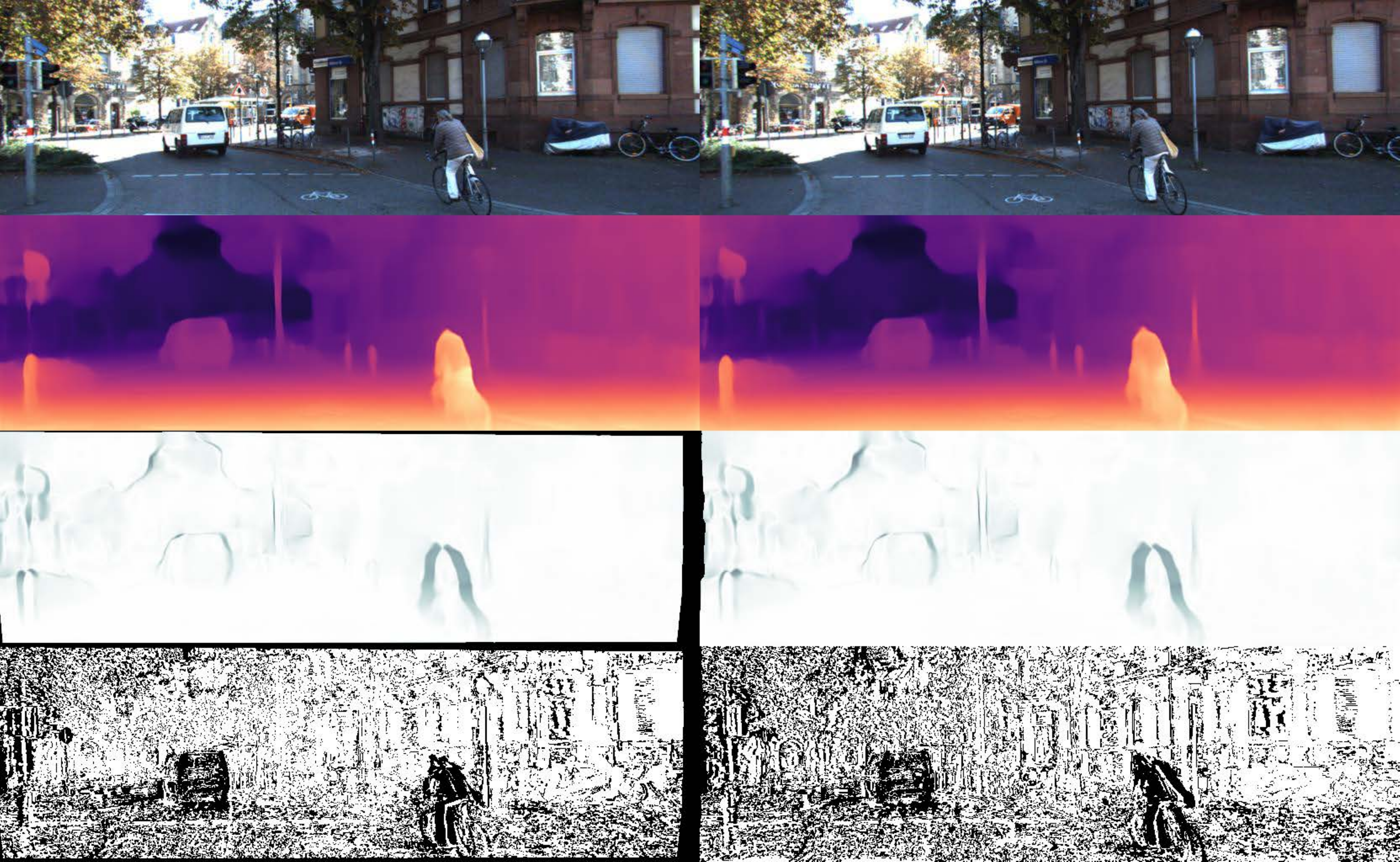}
  \addImg{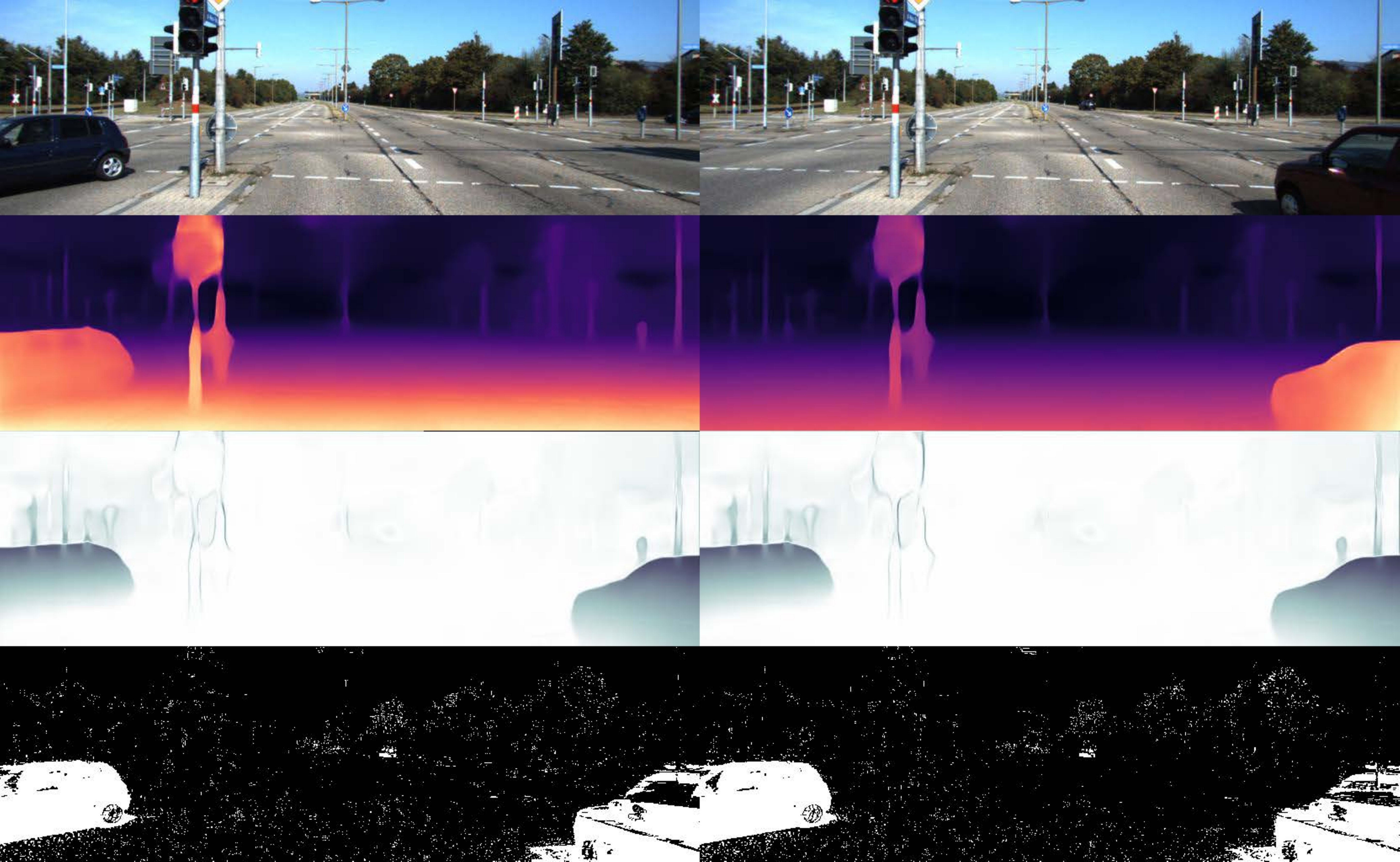}
  \caption{Visual results of depth and masking. 
  Top to bottom: sample image, estimated depth, 
  self-discovered mask $M_s$, and auto-mask $M_a$ \cite{monodepth2}.
  The proposed $M_s$ detects dynamics and occlusions (dark regions),
  and the binary mask $M_a$ finds invalid stationary points (black pixels).}
  \label{fig:show}
\end{figure*}

\subsection{Photometric and Smoothness Loss}\label{sec-photo-smooth}

Leveraging brightness constancy and spatial smoothness priors is ubiquitous 
in classical dense correspondence algorithms~\cite{baker2004lucas}.
Previous works~\cite{zhou2017unsupervised,yin2018geonet,ranjan2019cc} 
have used the photometric loss between the warped frame and 
the reference frame as an unsupervised loss function for network training.
With the predicted depth $D_a$ and pose $P_{ab}$, 
we synthesize $I'_a$ by warping $I_b$, 
where differentiable warping~\cite{jaderberg2015stn} 
is used.
With the synthesized $I'_a$ and the reference image $I_a$, 
we formulate the objective function as
\begin{equation}\label{eqn-photometric2}
L_{P}= \frac{1}{|\mathcal{V}|} \sum_{p \in \mathcal{V}} 
  (\lambda  \Vert I_a(p) - I'_a(p) \Vert _1 + 
  (1-\lambda)  \frac{1-\text{SSIM}_{aa'}(p)}{2} ),
\end{equation}
where $\mathcal{V}$ stands for the set of valid points that are successfully 
projected from $I_a$ to the image plane of $I_b$,
and $p$ stands for a generic point in $\mathcal{V}$.
We choose $L_1$ loss due to its robustness to outliers.
Besides, $\text{SSIM}_{aa'}$ stands for the element-wise similarity between 
$I_a$ and $I'_a$ by the SSIM function~\cite{wang2004image}.
This is used to better handle complex illumination changes since it normalizes the pixel illumination.
More specifically,
\begin{equation}
\text{SSIM (x,y)} = \frac{(2\mu_x\mu_y + C_1)(2\sigma_{xy} + C_2)}
    {(\mu_x^2 + \mu_y^2 + C_1)(\sigma_x^2 + \sigma_y^2 + C_2)},
\end{equation}
where $x,y$ stands for two $3$ by $3$ patches around the central pixel.
$C_1$ and $C_2$ are constants.
$\mu$ and $\sigma$ are local statistics of the image color, 
\ie, mean and variance, respectively.
Following~\cite{godard2017unsupervised, yin2018geonet, ranjan2019cc}, 
we use $C_1=0.0001$, $C_2=0.0009$, and $\lambda=0.15$.

As the photometric loss is not informative in low-texture regions of the scene,
existing work also incorporates a smoothness prior to 
regularize the estimated depth map.
We adopt the edge-aware smoothness loss used in~\cite{ranjan2019cc}.
Formally,
\begin{equation}
L_{S} = \sum_{p} ( e^{-\nabla I_a(p)} \cdot \nabla D_a(p) ) ^2,
\end{equation}
where $\nabla$ is the first derivative along spatial directions. 
It ensures smoothness to be guided by image edges.

\subsection{Geometry Consistency Loss}\label{sec-gc}

To explicitly enforce geometry consistency, we constrain that the predicted $D_a$ and $D_b$ (related by $P_{ab}$) 
conform the same 3D structure by penalizing their inconsistency.
Specifically, we propose a \emph{differentiable depth inconsistency} operation to compute the pixel-wise inconsistency between two depth maps, as shown in \figref{fig:gc-loss}.
Here, $D_b^{a}$ is the synthesized depth for $I_b$, which is generated by $D_a$ and pose $P_{ab}$ with the underlying rigid transformation.
$D'_b$ is an interpolation of $D_b$ for aligning and comparing with $D_b^{a}$.
Given them, we compute the depth inconsistency map $D_{\text{diff}}$ for each 
$p \in \mathcal{V}$ as:
\begin{equation}\label{eqn-depthdiff}
D_{\text{diff}}(p) = \frac{|D_b^{a}(p) - D'_b(p)|}{D_b^{a}(p) + D'_b(p)},
\end{equation}
%
where we normalize depth differences by their summation.
This works better than the absolute distance in practice as it treats points 
at different absolute depths equally in optimization.
Besides, the function is symmetric,
and the outputs are naturally ranging from $0$ to $1$, 
which makes the training more stable.

With the inconsistency map, 
we define the proposed geometry consistency loss as:
\begin{equation}\label{eqn-gc}
L_{G} = \frac{1}{|\mathcal{V}|} \sum_{p \in \mathcal{V}} D_{\text{diff}}(p),
\end{equation}
which minimizes the geometric inconsistency of predicted depths over two views.
By minimizing the depth inconsistency between samples in a batch, 
we naturally propagate such consistency to the entire sequence: 
the depth of $I_1$ agrees with the depth of $I_2$ in a batch; 
the depth of $I_2$ agrees with the depth of $I_3$ in another training batch. 
Eventually, depths of $I_i$ of a sequence should all agree with each other,
leading to scale-consistent results over the entire sequence.

\subsection{Masking Scheme}

The assumption of a moving camera and a static scene is underlying in the unsupervised depth learning framework,
where the moving objects in the scene and image pairs with identity camera pose provide invalid signals.
To be specific, the moving objects create the non-rigid flow that cannot be represented by the depth-based mapping, and the static camera consistently creates the identical flow that is independent to the depth prediction.
Therefore, we propose to mask out these regions by introducing a self-discovered mask ($M_s$) and adopting 
the auto-mask ($M_a$) by Monodepth2~\cite{monodepth2}.
The proposed $M_s$ computes weights (ranging from $0$ to $1$) for points in
$\mathcal{V}$ by checking their depth consistency,
and the $M_a$ simply removes invalid points from $\mathcal{V}$.
The proposed two masks are readily integrated into the proposed learning framework.

\subsubsection{Self Discovered Mask}\label{sec-mask}

As moving objects and occlusions naturally violate the geometry consistency assumption,
they will cause large depth inconsistency in our pre-computed $D_{\text{diff}}$ (\equref{eqn-depthdiff}).
This encourages us to define the $M_s$ as:
\begin{equation}\label{eqn-mask}
M_s = 1 - D_{\text{diff}},
\end{equation}
where the $M_s$ is in $[0, 1]$ and it attentively assign low weights for geometrically inconsistent pixels and high weights for consistent pixels.



\subsubsection{Auto-Mask}\label{sec-auto-mask}

To remove the invalid points in static pairs,
\eg, two images are captured at the same position,
we use the auto-mask $M_a$ that is proposed in \cite{monodepth2}.
It compares the photometric losses between the 
mapping by depth and pose and the identity mapping,
and it removes the points where the identity mapping leads to a lower loss.
Formally, for each $p \in \mathcal{V}$, we have
\begin{equation}\label{eqn-automask}
M_a(p) = 
  \begin{cases}
    1      &  \text{if  } \Vert I_a(p) - I'_a(p) \Vert _1 < \Vert I_a(p) - I_b(p) \Vert _1,  \\
    0      &  \text{otherwise}.
  \end{cases}
\end{equation}
Here $M_a$ is a binary mask for each point in $\mathcal{V}$ (valid points), 
and $I'_a$ is the warped image from the source image $I_b$ 
using the estimated depth and pose.
It makes the network to ignore objects that move at the same velocity as the camera, 
and it even ignores whole frames when the relative pose is identity.

\begin{figure*}[t]
  \centering
  \addImg[1]{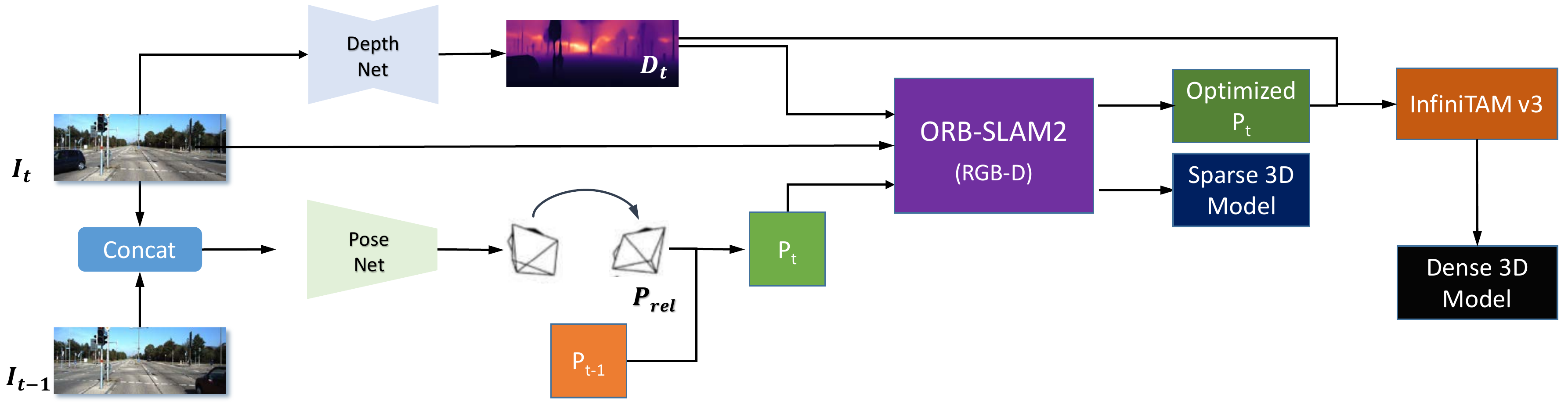}
  \caption{Pipeline of Pseudo-RGBD SLAM. 
  For the current frame $I_t$, 
  we first estimate its depth $D_t$ using our trained depth CNN.
  Then, we estimate its relative pose to previous frame $I_{t-1}$ (its pose 
  $P_{t-1}$ has been known in previous tracking) to recover the current pose.
  Next, we feed the color images, predicted depths, and estimated poses into ORB-SLAM2~\cite{murORB2},
  which outputs the accurate camera trajectory and a sparse 3D map. 
  Finally, given the consistent depth and camera trajectory, 
  we construct the dense voxel representation using InfiniTAMv3
  \cite{InfiniTAM_V3_Report_2017}.
  Note that the dense reconstruction here is only used for qualitative demonstration.
  }\label{fig:p-slam}
\end{figure*}

\subsubsection{How to use masks}\label{sec-use-mask}

First, to use $M_a$ in our loss function, 
we remove invalid points in $\mathcal{V}$ that have $M_a(p)=0$.
When training the network, we only compute losses on the remaining valid points.
Second, we use the proposed $M_s$ to re-weight the photometric loss 
in \equref{eqn-photometric2} by:
\begin{equation}\label{eqn-maskedphotometricloss}
L_{P}^M = \frac{1}{|\mathcal{V}|} \sum_{p \in \mathcal{V}} (M_s(p) \cdot L_{P} (p)). 
\end{equation}
This mitigates the noisy signals caused by moving objects and occlusions.
\figref{fig:show} shows visual results for the two types of masks, 
which coincides with our anticipation.
The dark regions in $M_s$ correspond to moving objects that 
violate the static scene assumption, 
\eg, the car region and human ride region.
In the binary $M_a$, black regions correspond to pixels that have similar speed with the camera,
\eg, the moving vehicle in the left example and the static scene in the right 
example.
\tabref{tab:ablation} shows the ablation study results, which shows that the proposed masks results in a significant performance improvement.

\section{Pseudo-RGBD SLAM}\label{sec-slam}

In this section, we present a Pseudo-RGBD SLAM system,
which is based on our trained models and existing SLAM systems.
We overview the system pipeline in \secref{sec-slam-overview},
followed by elaborating each component in \secref{sec-slam-details},
and finally, we discuss the advantages and limitations of the proposed system in \secref{sec-slam-discussion}.

\subsection{System Overview}\label{sec-slam-overview}

\figref{fig:p-slam} shows an overview of the proposed method,
which is composed of our SC-Depth, ORB-SLAM2~\cite{murORB2}, and InfiniTAMv3~\cite{InfiniTAM_V3_Report_2017}.
The whole system takes a monocular RGB video as input and outputs a globally consistent 6-DoF camera trajectory and sparse/dense 3D maps.
First, we initialize the tracking and mapping by using the predicted depth on the starting frame $I_0$,
which creates an initial 3D map.
Second, for a new frame $I_t$, we estimate its depth and relative pose to the previous view $I_{t-1}$ using our trained networks.  
As the camera pose of $I_{t-1}$ has been known from previous tracking or initialization, we can obtain the pose estimate for the current view by accumulation.
Third, we feed the color image, depth map, and the estimated pose for $I_t$ as input into ORB-SLAM2~\cite{murORB2},
which performs matching and optimization, 
resulting in an optimized camera pose as well as an increased map.
In such an incremental way, we eventually obtain a globally consistent camera trajectory and a sparse 3D map from the video.
Finally, we feed the color images, depth maps, and camera trajectories into InfiniTAMv3~\cite{InfiniTAM_V3_Report_2017},
which fuses depth maps to construct the dense and textured voxel volumes.

\subsection{System Details}\label{sec-slam-details}

\paragraph{ORB-SLAM2.}
The original RGB-D system takes the sensor captured depth as input,
while we use the estimated depth.
It relies on ORB features~\cite{rublee2011orb} to generate correspondences,
and it minimizes the reprojection error for pose optimization.
Poor correspondences (beyond the error threshold) are detected and removed as outliers, 
and the remaining correspondences are used for all sub-tasks, 
including tracking, mapping, loop closing, and re-localization.
We will elaborate on how our predicted depth and pose influence the correspondence and optimization in the system.


\paragraph{Depth.} 
The predicted depths are used to initialize a 3D map at the beginning, 
and they are used in the objective function during optimization.
Specifically, beyond 2d reprojection error, 
the system also minimizes the difference between the projected depth (from the 3D map to the image) and the predicted depth.
Formally,
\begin{equation}\label{eqn-reprojection}
    \begin{cases}
        E_{2D} = \sqrt{(p_x - p'_x)^2 + (p_y - p'_y)^2} \\ 
        E_{3D} = \sqrt{(p_x - p'_x)^2 + (p_y - p'_y)^2 + (p_d - p'_d)^2},
    \end{cases}
\end{equation}
where $p$ stands for points in current image plane,
and $p'$ stands for points projected from 3D map.
$p_d$ and $p'_d$ are their disparities, \ie, inverse depths.
Note that $p_d$ is computed from our predicted depth map, so it is unavailable in the monocular system.
This extends the reprojection error from 2D into 3D, 
which greatly improves the performance. 
Consequently, the consistency of estimated depths is vital in tracking.
For example, inconsistent depths would increase the reprojection error, 
and correct matches would be wrongly removed as outliers,
which causes the system to fail---See \figref{fig:keypoints}.

\paragraph{Pose.} 
The predicted pose is used as the initial pose during tracking,
in which the system first projects the sparse keypoints in a 3D map 
to the live view using the estimated pose and 
then searches for correspondences in the neighboring regions.
The camera pose is optimized through the Bundle Adjustment~\cite{murORB2}.
After tracking, we enrich the 3D map by unprojecting the keypoints detected in 
the live view to the map using the optimized camera pose.
The original ORB-SLAM2 uses the constant velocity motion model for 
initial pose, 
which simply assumes that the camera motion is the same as the previous frame.
Formally,
\begin{equation}\label{eqn-motion}
T_{t \rightarrow t+1} = 
    \begin{cases}
        T_{t-1 \rightarrow t} & \text{ORB-SLAM2}, \\ 
        \text{PoseNet}(I_t, I_{t+1}) & \text{Ours},
    \end{cases}
\end{equation}
where $T$ stands for relative pose.
However, this assumption is often violated in real scenarios, 
such as abrupt motion in driving scenes.
Though these frames are few in the sequence, 
they usually contribute the most of the drift in the final evaluation. 
Our trained pose CNN has the potential to cope with these cases.

\paragraph{InfiniTAMv3.}
It takes RGB-D videos and can densely reconstruct the scene structure.
We disable the internal tracking module and use our optimized 
camera poses and predicted depths for reconstruction.
This is only used for visualization purposes,
and it is also a demonstration of our consistent results.
Note that the dense reconstruction is very sensitive to geometry consistency,
\ie, it will crash when depths are not sufficiently consistent.
\figref{fig:demo} shows the screenshot of our demo,
which can be found in the supplementary material.

\subsection{Discussion}\label{sec-slam-discussion}

Our proposed SLAM system leverages the advantage of deep learning,
and it optimizes the predicted poses in the multi-view geometry-based framework.
This has distinctive advantages over existing solutions.

\paragraph{Advantages.}
Compared with classical monocular SLAM systems such as ORB-SLAM2~\cite{murORB2}, 
our advantages include:
\begin{enumerate}
  \item ORB-SLAM2 is often hard to initialize because it requires the multi-view triangulation,
  while our method can initialize at any time without latency by using the estimated dense depth---See \figref{fig:keypoints}.
  \item ORB-SLAM2 often loses tracking when the 3D map is over-sparse,
    while our method is more robust because we can enrich the map by using the predicted dense depth---See \figref{fig:keypoints}.
  \item  ORB-SLAM2 can only provide a sparse map,
  while our method enables dense 3D reconstruction by using the predicted dense depth---See \figref{fig:demo-kitti}. 
\end{enumerate}
Compared with learning-based methods~\cite{li2018undeepvo},
our advantage is the post geometric optimization \eg, Loop Closing~\cite{mur2014fast},
which can effectively correct drifts and improve the performance,
as shown in \figref{fig:vo}.

\paragraph{Limitations.}
Our method cannot recover the absolute scale because only monocular videos are used.
However, in real-world applications, the metric scale can be recovered by using other sensors and cues, like IMU and road landmarks.

\section{Experiments}

\subsection{Implementation details}\label{sec-implement-details}

\paragraph{Network architecture.}
Our depth network takes a single RGB image as input and outputs 
an inverse depth map.
It is a U-Net structure~\cite{ronneberger2015u}, 
and we use the ResNet50~\cite{He_2016_CVPR} encoder to extract features.
The decoder is the DispNet as used in~\cite{zhou2017unsupervised}.
The activations are sigmoids at the output layer and ELU nonlinearities~\cite{clevert2015fast} elsewhere. 
We convert the sigmoid output $x$ to depth with $D = 1/(ax + b)$, 
where $a$ and $b$ are chosen to constrain $D$ between 0.1 and 100 units.
It is a widely assumed depth range for outdoor driving scenes, 
which is the same with all related works~\cite{zhou2017unsupervised,yin2018geonet,ranjan2019cc}.
Besides, our pose network accepts two RGB frames as input and outputs their 6D relative camera pose.
We use the ResNet18~\cite{He_2016_CVPR} encoder to extract features.
In order to accept two frames, 
we modify the first layer to have six channels.
Then features are decoded to 6-DoF parameters via four convolutional layers.

\paragraph{Single scale supervision.}
Previous methods compute the losses on an image pyramid, \ie, usually four layers.
They either work on the decoder's side outputs~\cite{zhou2017unsupervised,yin2018geonet,zou2018df} 
or upsample them to the original image resolution~\cite{monodepth2}.
However, it introduces great computational overhead in training.
By contrast, we only compute the loss on the original image resolution.
This has a less computational cost and achieves on par performance with the multi-scale solution in MonoDepth2~\cite{monodepth2}, as shown in \tabref{tab:ablation}.
The motivation is that we empirically find that the supervision  on low-resolution images is inaccurate,
and the camera movement between training image pairs is small so that the multi-scale solution is unnecessary.

\paragraph{Training details.}
We implement the proposed method using the PyTorch~\cite{paszke2017automatic}.
Following~\cite{zhou2017unsupervised,ranjan2019cc,Wang2018CVPR}, 
we use a snippet of three sequential video frames as a training sample.
We compute the projection and losses from the second frame to others 
and reverse them again for maximizing the data usage.
The images are augmented with random scaling, cropping, and horizontal flips during training.
We use ADAM~\cite{kingma2014adam} optimizer and set the learning rate to be $10^{-4}$.
During training, we set $\alpha = 1.0$, $\beta = 0.1$, and $\gamma = 0.5$ in~\equref{eqn-totalloss}.
For fast convergence, we initialize the encoder of our networks by using the pre-trained model on ImageNet~\cite{imagenet_cvpr09}.

\begin{table*}[t]
  \setlength{\tabcolsep}{1.0mm}
  \caption{Single-view depth estimation results on KITTI~\cite{Geiger2013IJRR}. Legends: D---depth supervision; S---stereo pairs; M---monocular snippets; L---semantic labels or networks; F---joint learning with optical flow. 
  }\label{tab:depth}
  \centering
  \begin{tabular}{l c c| c c c c | c c c}
    \hline
     & & & \multicolumn{4}{c|}{Error $\downarrow$} & \multicolumn{3}{c}{Accuracy $\uparrow$}  \\
     \cline{4-10}
     Methods & Resolution & Supervision & AbsRel & SqRel & RMS & RMSlog & $\delta_1$ & $\delta_2$ & $\delta_3$ \\
     \hline
     \cite{eigen2014depth} & $612 \times 184$ &  D & 0.203 & 1.548 & 6.307 & 0.282 & 0.702 & 0.890 & 0.958 \\
     \cite{kuznietsov2017semi} & $621 \times 187$ & S+D & 0.113 & 0.741 & 4.621 & 0.189 & 0.862 & 0.960 & 0.986 \\
     DORN~\cite{fu2018deep} & $513 \times 385$ & D & \textbf{0.072} & \textbf{0.307} & 2.727 & 0.120 & 0.932 & 0.984 & 0.994 \\
     VNL~\cite{Yin2019enforcing} & $385 \times 385$ & D & \textbf{0.072} & - & 3.258 & \textbf{0.117} & \textbf{0.938} & \textbf{0.990} & \textbf{0.998} \\
     \hline
     \cite{garg2016unsupervised} & $620 \times 188$ &S & 0.152 & 1.226 & 5.849 & 0.246 & 0.784 & 0.921 & 0.967 \\
     \cite{godard2017unsupervised} & $512 \times 256$ & S & 0.148 & 1.344 & 5.927 & 0.247 & 0.803 & 0.922 & 0.964 \\
     \cite{zhan2018unsupervised} & $608 \times 160$ & S+M & 0.144 & 1.391 & 5.869 & 0.241 & 0.803 & 0.928 & 0.969 \\
     SuperDepth+pp~\cite{pillai2019superdepth} & $1024 \times 382$ & S & 0.112 & 0.875 & 4.958 & 0.207 & 0.852 & 0.947 & 0.977 \\
     Monodepth2-S\cite{monodepth2} & $1024 \times 320$ & S & 0.107 & 0.849 & 4.764 & 0.201 & 0.874 & 0.953 & 0.977 \\
     Monodepth2-MS\cite{monodepth2} & $1024 \times 320$ & S+M & 0.106 & 0.806 & 4.630 & 0.193 & 0.876 & 0.958 & 0.980 \\
     D3VO~\cite{yang2020d3vo} & $512 \times 256$ & S+M & \textbf{0.099} & \textbf{0.763} & \textbf{4.485} & \textbf{0.185} & \textbf{0.885} & \textbf{0.958} & \textbf{0.979} \\
     \hline
     Geonet-Resnet~\cite{yin2018geonet} & $416 \times 128$ & M+F & 0.155 & 1.296 & 5.857 & 0.233 & 0.793 & 0.931 & 0.973\\
     DF-Net~\cite{zou2018df} & $576 \times 160$ & M+F & 0.150 & 1.124 & 5.507 & 0.223 & 0.806 & 0.933 & 0.973 \\
     Struct2Depth~\cite{casser2019struct2depth} & $416 \times 128$ & M+L & 0.141 & 1.026 & 5.291 & 0.215 & 0.816 & 0.945 & 0.979 \\
     DW\cite{gordon2019depth} & $416 \times 128$ & M+L & 0.128 & 0.959 & 5.230 & 0.212 & 0.845 & 0.947 & 0.976 \\
     GLNet\cite{chen2019self} & $416 \times 128$ & M+F & 0.135 & 1.070 & 5.230 & 0.210 & 0.841 & 0.948 & 0.980 \\
     CC~\cite{ranjan2019cc} & $832 \times 256$ & M+F & 0.140 & 1.070 & 5.326  & 0.217 & 0.826 & 0.941 & 0.975 \\
     EPC++~\cite{luo2019every} & $832 \times 256$ &M+F & 0.141 & 1.029 & 5.350 & 0.216 & 0.816 & 0.941 & 0.976 \\
     \cite{zhao2020towards} & $832 \times 256$ &M+F & 0.113 & \textbf{0.704} & 4.581 & 0.184 & 0.871 & 0.961 & \textbf{0.984} \\
     Insta-DM~\cite{lee2021learning} & $832 \times 256$ & M+F+L & 0.112	& 0.777	& 4.772	& 0.191	& 0.872	& 0.959	& 0.982 \\
     SGDepth-full~\cite{klingner2020self} & $1280 \times 384$ & M+L & 0.107 & 0.768 & 4.468 & 0.186 & 0.891 & 0.963 & 0.982 \\
     PackNet-Sem~\cite{packnet-semguided} & $1280 \times 384$ & M+L & \textbf{0.100} & 0.761 & \textbf{4.270} & \textbf{0.175} & \textbf{0.902} & \textbf{0.965} & 0.982 \\
     \hline
     SfMLearner~\cite{zhou2017unsupervised} & $416 \times 128$ & M & 0.208 & 1.768 & 6.856 & 0.283 & 0.678 & 0.885 & 0.957 \\
     \cite{yang2018unsupervised} & $416 \times 128$ & M & 0.182 & 1.481 & 6.501 & 0.267 & 0.725 & 0.906 & 0.963 \\
     Vid2Depth~\cite{mahjourian2018unsupervised} & $416 \times 128$ & M & 0.163 & 1.240 & 6.220 & 0.250 & 0.762 & 0.916 & 0.968 \\
     DDVO~\cite{Wang2018CVPR} & $416 \times 128$ & M & 0.151 & 1.257 & 5.583 & 0.228 & 0.810 & 0.936 & 0.974 \\
     \cite{Zhou_2019_ICCV} & $1248 \times 384$ & M & 0.121 & 0.837 & 4.945  & 0.197 & 0.853 & 0.955 & 0.982 \\
     MonoDepth2~\cite{monodepth2} & $1024 \times 320$ & M & 0.115 & 0.882 & 4.701 & 0.190 & 0.879 & 0.961 & 0.982 \\
     \cite{li2020unsupervised} & $416 \times 128$ & M & 0.130 & 0.950 & 5.138 & 0.209 & 0.843 & 0.948 & 0.978 \\
     PackNet-SfM~\cite{packnet} & $1280 \times 384$ & M & \textbf{0.107} & \textbf{0.802} & \textbf{4.538} & \textbf{0.186} & \textbf{0.889} & \textbf{0.962} & \textbf{0.981} \\
     \hline
     Ours-R18 & $416 \times 128$ & M &   0.132  &   0.982  &   5.226  &   0.209  &   0.835  &   0.947  &   0.978  \\
     Ours-R50 & $416 \times 128$ & M &   0.126  &   0.920  &   5.245  &   0.208  &   0.840  &   0.949  &   0.979  \\ 
     Ours-R18 & $832 \times 256$ & M &   0.119  &   0.857  &   4.950  &   0.197  &   0.863  &   0.957  &   0.981  \\
     Ours-R50 & $832 \times 256$ & M &  \textbf{0.114} & \textbf{0.813} & \textbf{4.706} & \textbf{0.191} & \textbf{0.873} & \textbf{0.960} & \textbf{0.982} \\
     \hline
  \end{tabular}
\end{table*}

\begin{table}[t]
  \setlength{\tabcolsep}{1mm}
  \caption{Ablation study results on KITTI. We use the ResNet18 model, and the image resolution is $416 \times 128$.
  }\label{tab:ablation}
  \centering
  \begin{tabular}{l|c c c }
    \hline
    Models & AbsRel & $\delta_1$ & Time \\
    \hline
    Baseline (B)  & 0.200 & 0.786 & \multirow{5}{*}{10h} \\
    B+$L_{G}$  & 0.155 & 0.786 & \\
    B+$L_{G}$+$M_s$ & 0.137 & 0.822 &  \\
    B+$L_{G}$+$M_a$ & 0.153 & 0.797 &  \\
    B+$L_{G}$+$M_s$+$M_a$ (Ours) & \textbf{0.132} & \textbf{0.835} &  \\
    \hline
    Ours with NCC & 0.145 & 0.804 & 11h \\
    Ours + Multi-Scale & 0.134 & 0.829 & 21h \\
    \hline
  \end{tabular}
\end{table}

\begin{table}[t]
  \setlength{\tabcolsep}{1mm}
  \caption{Trade-offs between image resolution, network, and speed.
    We train models on KITTI using a TESLA V100 GPU and test the inference 
    speed in an RTX 2080 GPU.
  }\label{tab:network-resolution}
  \centering
  \begin{tabular}{l c |c c c c} \hline
    Resolution & Model & AbsRel & $\delta_1$ & Train  & Infer \\ \hline
    \multirow{2}{*}{$416 \times 128$ } & R18 & 0.132 & 0.835 & \textbf{10h} & \textbf{228 fps} \\
    & R50 & 0.126 & 0.840 & 16h & 110 fps \\ \cline{1-2}
    \multirow{2}{*}{$832 \times 256$} & R18 & 0.119 & 0.863 & 29h & 133 fps \\
    & R50 & \textbf{0.114} & \textbf{0.873} &  37h & 59 fps\\ \hline
  \end{tabular}
\end{table}

\begin{table*}[t]
\centering
    \caption{Single-view depth estimation results on NYUv2~\cite{silberman2012indoor}. Legends: D---depth supervision; M---unsupervised training using monocular snippets; F---joint learning with the optical flow;  WR---weak rectification~\cite{bian2020unsupervised} which pre-processes the hand-held camera captured videos for better training. More specifically, \cite{bian2020unsupervised} remove the relative rotation between training pairs since they find that it is hard for the pose network to learn image rotation.
    }
    \label{tab:nyu}
     \setlength{\tabcolsep}{2mm}
    \begin{tabular}{l c c | c c c  | c c c}
     \hline
     \multirow{2}{*}{Methods} & \multirow{2}{*}{Resolution} & \multirow{2}{*}{Supervision} & \multicolumn{3}{c|}{Error $\downarrow$} & \multicolumn{3}{c}{Accuracy $\uparrow$}  \\
     \cline{4-9}
      & & & AbsRel & Log10 & RMS  & $\delta_1$ & $\delta_2$ & $\delta_3$ \\
     \hline
     Make3D~\cite{saxena2006learning} & - & D &  0.349 & - & 1.214 & 0.447 & 0.745 & 0.897 \\
     \cite{wang2015towards} & - & D & 0.220 & 0.094 & 0.745 & 0.605 & 0.890 & 0.970 \\
     \cite{eigen2015predicting} & $320 \times 240$ & D & 0.158 & - & 0.641 & 0.769 & 0.950 & 0.988 \\
     \cite{chakrabarti2016depth} & $561 \times 427$ & D & 0.149 & - & 0.620 & 0.806 & 0.958 & 0.987 \\
     \cite{laina2016deeper} & $304 \times 228$ & D & 0.127 & 0.055 & 0.573 & 0.811 & 0.953 & 0.988 \\
     \cite{li2017two} & $310 \times 232$  & D & 0.143 & 0.063 & 0.635 & 0.788 & 0.958 & 0.991 \\
     DORN~\cite{fu2018deep} & $353 \times 257$ & D & 0.115 & 0.051 & 0.509 & 0.828 & 0.965 & 0.992 \\
     VNL~\cite{Yin2019enforcing} & $385 \times 385$ & D & \textbf{0.108} & \textbf{0.048} & \textbf{0.416} & \textbf{0.875} & \textbf{0.976} & \textbf{0.994} \\
     \hline
     \cite{Zhou_2019_ICCV} & $256 \times 192$ & M+F & 0.208 & 0.086 & 0.712 & 0.674 & 0.900 & 0.968 \\
     \cite{zhao2020towards} & $576 \times 448$ & M+F & \textbf{0.189} & \textbf{0.079} & \textbf{0.686} & \textbf{0.701} & \textbf{0.912} & \textbf{0.978} \\
     \hline
     Monodepth2~\cite{monodepth2} & $320 \times 256$ & M & 0.176  &   0.074  &   0.639  &   0.734  &   0.937  &   0.983 \\ 
     Ours-R18 & $320 \times 256$ & M  &  0.159  &   0.068  &   0.608  &   0.772  &   0.939  &   0.982  \\
     Ours-R50 & $320 \times 256$ & M  & \textbf{0.157}  &   \textbf{0.067}  &   \textbf{0.593}  &   \textbf{0.780}  &   \textbf{0.940}  &   \textbf{0.984}  \\
     \hline
     Monodepth2~\cite{monodepth2} & $320 \times 256$ & WR-M & 0.151  &  0.064  &   0.559  &   0.795  &  0.947 &   0.985 \\ 
     Ours-R18 & $320 \times 256$ & WR-M & 0.143  &   \textbf{0.060}  &   0.538  &   0.812  &   0.951  &   0.986   \\
     Ours-R50 & $320 \times 256$ & WR-M & \textbf{0.142}  &   \textbf{0.060}  &   \textbf{0.529}  &   \textbf{0.813}  &   \textbf{0.952}  &   \textbf{0.987}  \\
     \hline
    \end{tabular}
\end{table*}

\paragraph{Datasets.}
For depth estimation evaluation, we use both the KITTI~\cite{Geiger2013IJRR} and NYUv2~\cite{silberman2012indoor} datasets.
In KITTI, we use the same training/testing split as in~\cite{zhou2017unsupervised}.
The 697 images are used for testing.
We train the network for $200K$ iterations, where we set the batch size to be $4$ and resize images to $832 \times 256$ resolution for training.
In NYUv2, we use the officially provided 654 densely labeled images for testing,
and use the rest sequences (no overlapping with the testing scenes) for training.
We extract one frame from every $10$ frame in the original video to remove redundant frames,
and we resize images to $320 \times 256$ resolution as input of the network.
We train models for $50$ epochs, and the batch size is $8$.
For visual odometry evaluation,
we use the KITTI odometry dataset (Seq. 00-08) for training, and we test our method on the Seq. 09-10.
Moreover, we use the KAIST urban dataset~\cite{jeong2019complex} to validate the zero-shot generalization ability of our method.
We use one of the hardest scenes (urban39-pankyo), which contains more than $18000$ street-view images,
and we split it into $9$ sequences with each sequence containing $2000$ images for testing.

\begin{figure}[t]
  \centering
  \addImg[0.48]{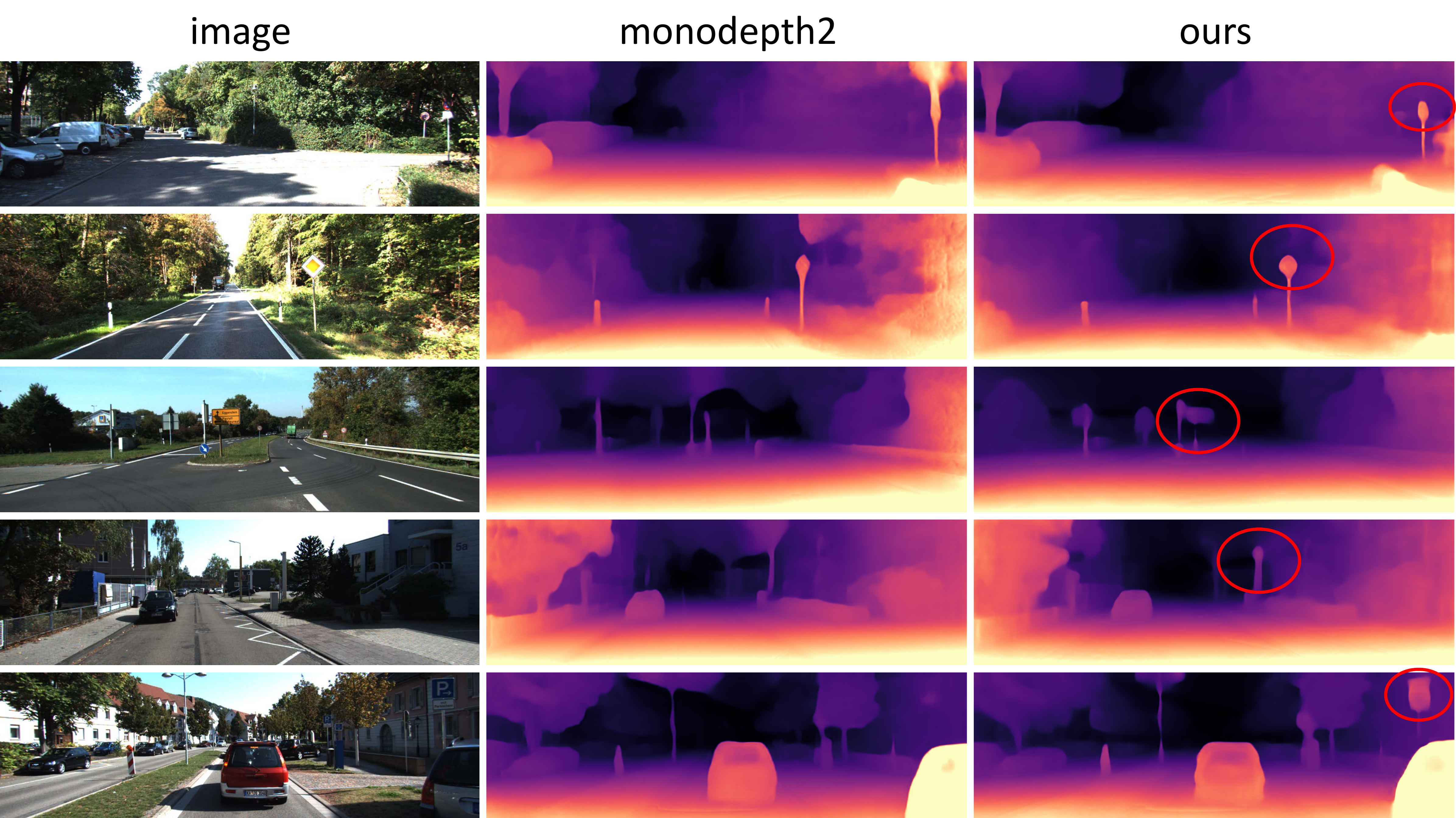}
  \caption{Qualitative comparison with the Monodepth2~\cite{monodepth2} on KITTI.
  }\label{fig:kitti_vis}
\end{figure}


\paragraph{Evaluation metrics.}
For depth evaluation, following previous methods~\cite{Yin2019enforcing, zhou2017unsupervised}, we use the mean absolute relative error (AbsRel),
mean log10 error (Log10), root mean squared error (RMS), root mean squared log error (RMSlog), 
and the accuracy under threshold ($\delta_i$ $<$ $1.25^i$, $i = 1, 2, 3$).
As unsupervised methods cannot recover the absolute scale, we multiply the predicted depth maps by a scalar that matches the median with that of the ground truth,
as in~\cite{zhou2017unsupervised}.
The predicted depths are capped at $80m/10m$ in KITTI and NYUv2 datasets, respectively.
For visual odometry evaluation, we follow the standard evaluation metrics,
including the translational ($t_{err}$) and rotational errors ($r_{err}$) averaged over the entire sequence~\cite{Geiger2013IJRR},
and the absolute trajectory error (ATE)~\cite{sturm12iros}.

\subsection{Depth Estimation}

\begin{figure}[ht]
\centering
\includegraphics[width=0.95\linewidth]{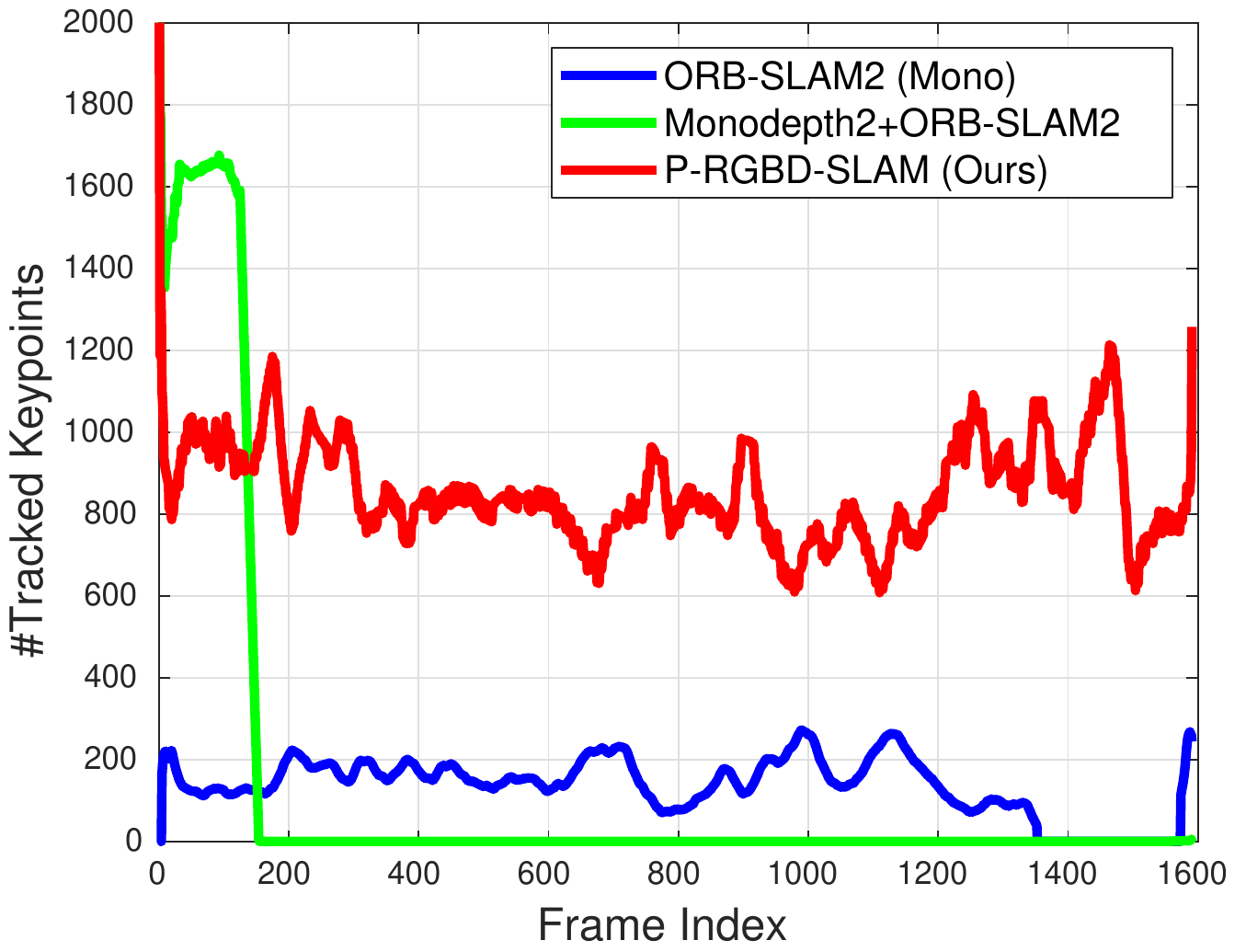}
\caption{Number of tracked keypoints on Seq. 09. We extract $2000$ feature points for all methods, 
and the values in the figure are smoothed for visualization.
}
\label{fig:keypoints}
\end{figure}

\paragraph{Results on KITTI.}
\tabref{tab:depth} shows the results,
which shows that the supervised methods~\cite{fu2018deep, Yin2019enforcing} are best-performing,
followed by the stereo trained models~\cite{yang2020d3vo}.
Besides, it shows that learning with semantic labels~\cite{packnet-semguided} or optical flow~\cite{zhao2020towards} can effectively improve the performance of monocular methods.
We are here more interested in the monocular methods that do not use additional information.
In this category, our method outperforms previous methods (before 2020),
and it shows on par performance with the MonoDepth2~\cite{monodepth2}.
However, we argue that our advantage against Monodepth2 is the depth consistency (\tabref{tab:consistency}),
which has important implications on downstream video-based tasks.
For example, contributed to the consistent depth prediction, our method can be readily plugged into the Visual SLAM systems,
while the Monodepth2 is unable---See \figref{fig:keypoints} for detailed analysis.

\paragraph{Efficacy of the proposed methods.}
\tabref{tab:ablation} summarizes the results.
It shows that the proposed $L_{G}$ makes training more stable by enforcing depth consistency,
and the proposed $M_s$ can boost performance significantly by handling scene dynamics.
Besides, it shows that using $M_a$ can contribute to extra marginal performance improvement by removing the stationary points.
Consequently, the final solution (with all terms) can achieve the best performance.
Moreover, \tabref{tab:network-resolution} shows the relation between depth accuracy, training time, inference speed, network architecture, and image resolution.

\begin{figure}[t]
  \centering
  \addImg[0.48]{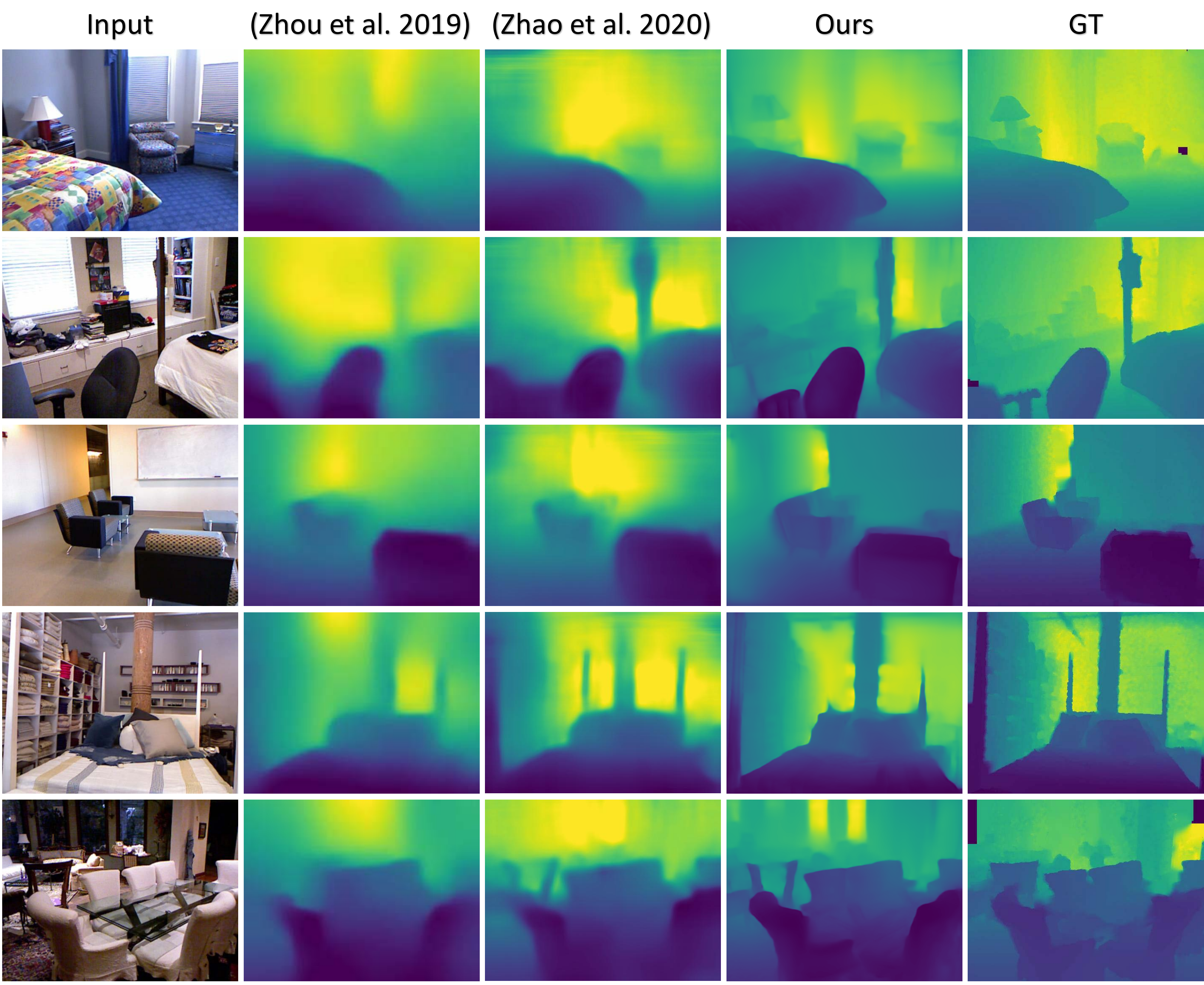}
  \caption{Qualitative comparison with the state-of-the art unsupervised methods on NYUv2.}
  \label{fig:nyu-vis}
\end{figure}

\begin{table*}[t]
\centering
    \setlength{\tabcolsep}{0.1mm}
    \caption{Visual odometry results on KITTI~\cite{Geiger2013IJRR}. S/M stands for training on stereo/monocular videos, and G stands for geometric optimization. \xmark ~stands for failure in initialization or tracking.
    }
    \label{tab:vo}
    \begin{tabular}{l c | c c c | c c c}
     \hline
     \multirow{2}{*}{Methods} & \multirow{2}{*}{Types} & \multicolumn{3}{c|}{Seq. 09} & \multicolumn{3}{c}{Seq. 10} \\
     & & $t_{err}$ ($\%$) & $r_{err}$ ($^{\circ}/100m$) & ATE(m) & $t_{err}$ ($\%$) & $r_{err}$ ($^{\circ}/100m$) & ATE(m) \\
     \hline
    Depth-VO-Feat~\cite{zhan2018unsupervised} & S & 11.89 & 3.60 & \textbf{52.12} & 12.82 & 3.41 & \textbf{24.70} \\
    UndeepVO~\cite{li2018undeepvo} & S & 7.01 & 3.60 & - & 10.63 & 4.60 & - \\
    PoseGraph~\cite{li2019pose} & S + G & 6.23 & 2.11  & - & 12.9 & 3.17 & - \\ 
    DVSO~\cite{yang2018unsupervised} & S + G & 0.83 & \textbf{0.21} & - & 0.74 & \textbf{0.21} & - \\
    D3VO~\cite{yang2020d3vo} & S + G & \textbf{0.78} & - & - & \textbf{0.62} & - & - \\
    \hline
    SfMLearner~\cite{zhou2017unsupervised} & M & 19.15 & 6.82 & 77.79 & 40.40 & 17.69 & 67.34 \\ 
    GeoNet~\cite{yin2018geonet} & M &  28.72 & 9.8 & 158.45 & 23.90 & 9.0 & 43.04 \\
    DeepMatchVO~\cite{shen2019beyond} & M & 9.91 & 3.8 & 27.08 & 12.18 & 5.9 & 24.44 \\
    MonoDepth2~\cite{monodepth2} & M & 17.17 & 3.85 & 76.22 & 11.68 & 5.31 & 20.35 \\
    DW~\cite{gordon2019depth}-Learned & M & - & - & 20.91 & - & - & 17.88 \\
    DW~\cite{gordon2019depth}-Corrected & M & - & - & 19.01 & - & - & 14.85 \\
    \cite{zou2020learning} & M & \textbf{3.49} & 1.00 & \textbf{11.30} & 5.81 & 1.8 & \textbf{11.80} \\
    \cite{zhao2020towards} & M + G & 6.93 & \textbf{0.44} & - & \textbf{4.66} & \textbf{0.62} & - \\
    \hline
    SC-Depth (Ours w/o $L_G$) & M &  12.43 & 4.65 & 83.27 & 11.86 & 4.95 & 21.19 \\
    SC-Depth (Ours) & M &  7.31 & 3.05 & 23.56 & 7.79 & 4.90 & 12.00 \\
     Pseudo-RGBD SLAM (MonoDepth2) & M + G & \xmark & \xmark & \xmark & \xmark & \xmark & \xmark \\
     Pseudo-RGBD SLAM (Ours w/o $L_G$) & M + G & 7.81 & 2.44 & 34.15 & 7.62 & 2.41 & 9.02 \\
     Pseudo-RGBD SLAM (Ours - Motion Model) & M + G & 5.70 & 1.33 & \textbf{13.22} & \textbf{3.82} & \textbf{1.76} & \textbf{5.96} \\
     Pseudo-RGBD SLAM (Ours - Pose CNN) & M + G & \textbf{5.08} & \textbf{1.05} & 13.40 & 4.32 & 2.34 & 7.99 \\
     \hline
    \end{tabular}
\end{table*}

\begin{figure*}[t]
\centering
\includegraphics[width=0.45\linewidth]{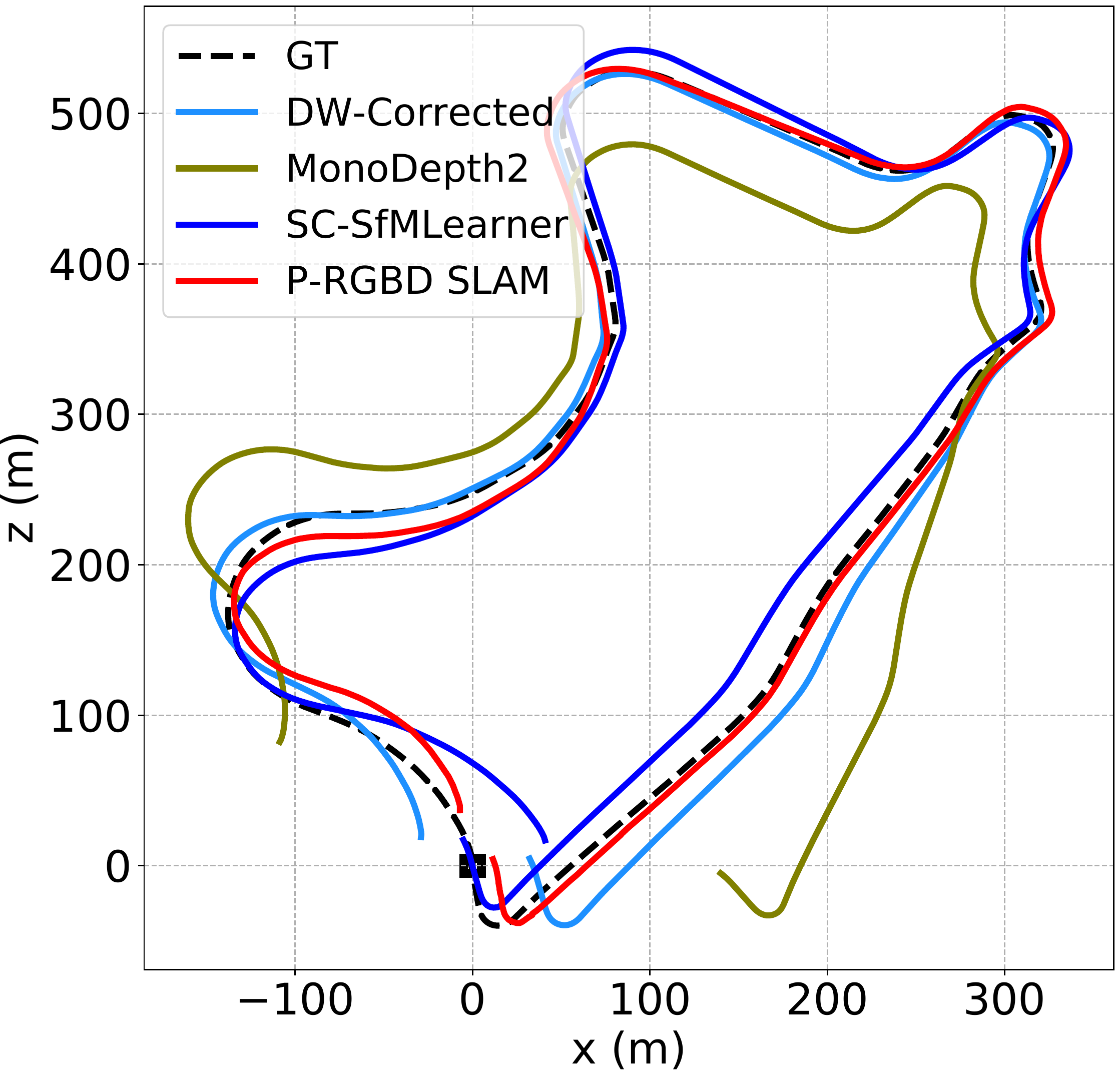}
\includegraphics[width=0.45\linewidth]{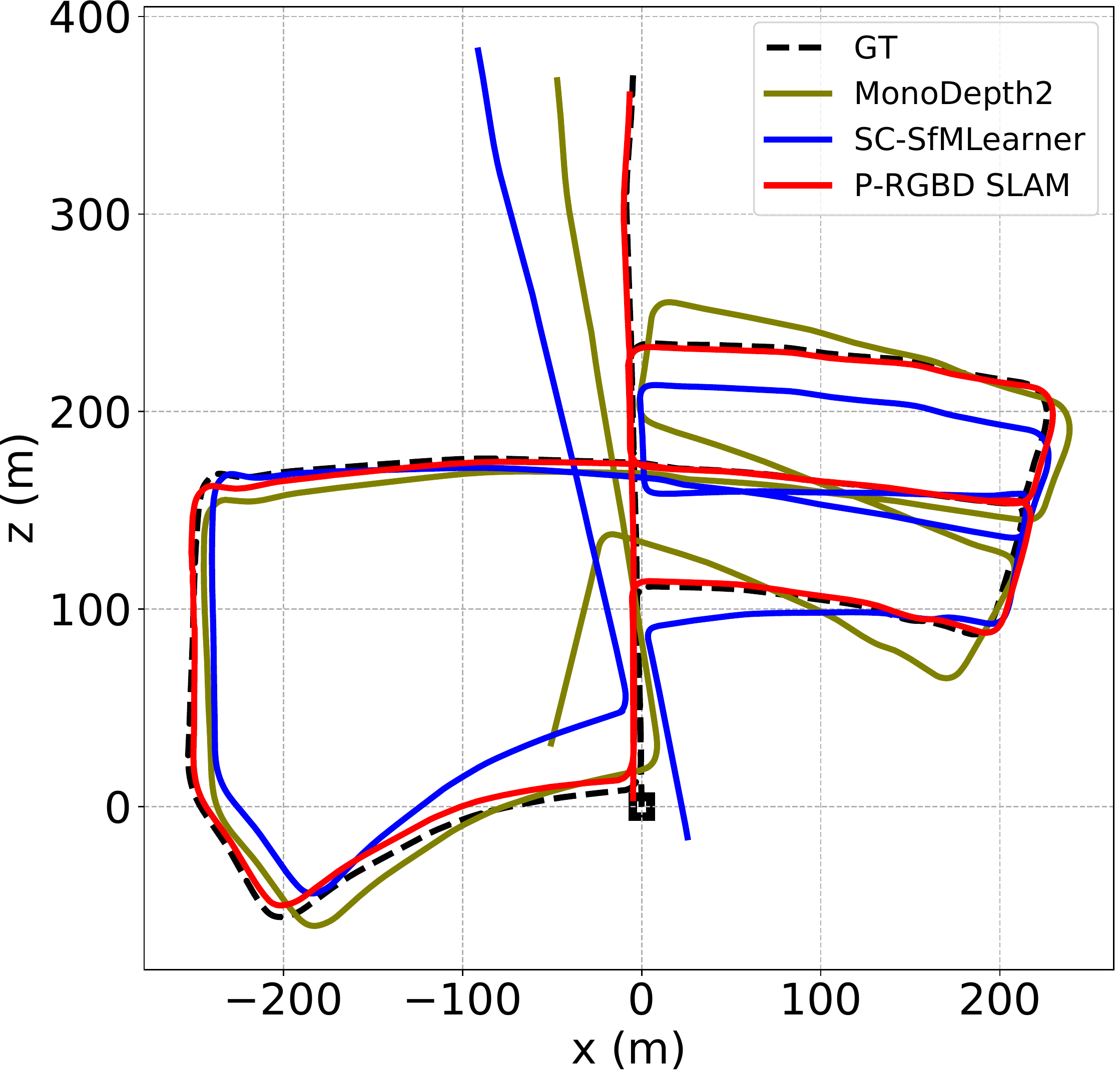}
\caption{Estimated trajectory on Seq. 09 (left) and 05 (right). 
The results optimized by the proposed Pseudo-RGBD SLAM are more accurate than our SC-Depth and other learning-based methods,
and the improvement is especially large when loops are detected and closed. For example, the $t_{err}$ is reduced from $5.91$ to $1.67$ on Seq. 05.
}
\label{fig:vo}
\end{figure*}

\paragraph{Multi-scale supervision.}
\tabref{tab:ablation} shows the results of our method with the modified multi-scale solution
proposed in~\cite{monodepth2}.
It upsamples the predicted four depth maps to original image resolution and
then computes losses instead of downsampling the original color image
\cite{zhou2017unsupervised}.
The result demonstrates that our method could hardly benefit from that,
and it requires two times longer time for training.
Therefore, we use single-scale supervision in our framework.

\paragraph{SSIM vs NCC.}
\tabref{tab:ablation} shows the results of our method with the normalized cross-correlation (NCC) loss,
in which we replace the SSIM.
Both losses compute the local image similarity on a 3 by 3 patch.
The results show that SSIM leads to better performance than NCC in our unsupervised learning framework.

\paragraph{Results on NYUv2.}
\tabref{tab:nyu} shows the results, 
which shows that our method outperforms previous unsupervised methods by a large margin.
Besides, following \cite{bian2020unsupervised}, we remove the relative rotation between training image pairs since they find that it is hard for the pose network to learn image rotation.
This leads to a significant improvement because rotation is the dominate ego-motion in hand-held camera captured videos.
\cite{zhao2020towards} solves the problem by replacing the Pose CNN with a traditional geometry-based pose solver.
The qualitative results are shown in \figref{fig:nyu-vis}.
We find that MonoDepth2~\cite{monodepth2} often collapses in training,
so we report the best result.
Compared with the supervised methods, our method is inferior to the state-of-the-art~\cite{Yin2019enforcing} but outperforms many previous methods~\cite{liu2016learning, saxena2006learning,  wang2015towards, eigen2015predicting, chakrabarti2016depth, li2017two}.


\begin{table}[t]
\centering
    \setlength{\tabcolsep}{1.0mm}
    \caption{Depth consistency results on Seq. 09. Fitness measures the overlapping area of two point clouds (\# of inlier correspondences / \# of points in target).
    RMSE is averaged over all inlier correspondences (\#Corr).}
    \label{tab:consistency}
    \begin{tabular}{l | c c c }
     \hline
     Methods & Fitness ($\uparrow$) & RMSE ($\downarrow$) & \#Corr ($\uparrow$) \\
     \hline
    MonoDepth2 & 0.384 & 9.84e-3 & 80.776K \\
    Ours (w/o $L_G$) &  0.663 & 8.90e-3 & 129.825K \\  
    Ours & \textbf{0.689} & \textbf{8.71e-3} & \textbf{134.956K} \\  
     \hline
    \end{tabular}
\end{table}

\begin{table}[t]
  \setlength{\tabcolsep}{0.5mm}
  \caption{Visual odometry results on KITTI. We evaluate the results on all frames and on keyframes that are selected by the ORB-SLAM2 since the latter cannot provide results for the full sequence due to unsuccessful initialization or tracking failure. The ATE (m) metric is used. We use $2K$ keypoints as default, and we analyze the effect of keypoint numbers on system performance by increasing it to $8K$. 
  }\label{tab:vo_orb}
  \centering
  \begin{tabular}{| l | c | c | c | c | c | c |}
    \cline{2-7}
    \multicolumn{1}{c|}{} & \multicolumn{3}{c|}{ORB KeyFrames} & \multicolumn{3}{c|}{All Frames} \\
    \hline
    Seq & Frames & ORB & Ours & Frames & Ours-2K & Ours-8K \\
    \hline
    00 & 1928 & \textbf{6.33} & 6.43 & 4541 & \textbf{6.05} & 6.57 \\
    \hline
    01 & 395 & 468.9 & \textbf{299.11} & 1101 & \textbf{289.29} & 301.08 \\
    \hline
    02 & 2445 & 26.18 & \textbf{8.86}  & 4661 & \textbf{8.79} & 9.42 \\
    \hline
    03 & 361 & \textbf{1.21} & 2.92 & 801 & \textbf{3.00} & 3.71 \\
    \hline
    04 & 149 & \textbf{1.73} & 2.75 & 271 & \textbf{2.64} & 2.89 \\
    \hline
    05 & 1129 & 4.78 & \textbf{4.47} & 2761 & \textbf{4.35} & 5.58 \\
    \hline
    06 & 479 & 13.34 & \textbf{4.11} & 1101 & \textbf{4.35} & 6.10 \\
    \hline    
    07 & 467 & 2.28 & \textbf{0.77} & 1101 & \textbf{0.76} & 2.39 \\
    \hline    
    08 & 2129 & 49.23 & \textbf{18.48} & 4071 & \textbf{19.37} & 20.10 \\
    \hline
    09 & 871 & 50.78 & \textbf{13.43} & 1591 & \textbf{13.40} & 17.16 \\
    \hline    
    10 & 549 & \textbf{7.26} & 7.52 & 1201 & 7.99 & \textbf{6.48}\\
    \hline
  \end{tabular}
\end{table}

\begin{table}[t]
  \setlength{\tabcolsep}{1mm}
  \caption{Zero-short generalization on KAIST dataset~\cite{jeong2019complex}. We compare our method with ORB-SLAM2 using the ATE (m) metric. 
  }\label{tab:kaist}
  \centering
  \begin{tabular}{| l | c | c | c | c |}
    \cline{2-5}
    \multicolumn{1}{c|}{} & \multicolumn{3}{c|}{ORB KeyFrames} & \multicolumn{1}{c|}{All Frames (2K)} \\
    \hline
    Seq & Frames & ORB & Ours  & Ours \\
    \hline
    00 & 189 & 7.04 & \textbf{2.35} & 2.21 \\
    \hline
    01 & 286 & 21.06 & \textbf{3.90} & 4.29 \\
    \hline
    02 & 231 & 11.95 & \textbf{4.65} & 5.11 \\
    \hline
    03 & 150 & 11.67 & \textbf{2.71} & 2.59 \\
    \hline
    04 & 140 & 3.80 & \textbf{2.00}  & 1.67 \\
    \hline
    05 & 201 & 55.87 & \textbf{27.46} & 28.34 \\
    \hline
    06 & 306 & 136.85 & \textbf{7.47} & 7.78 \\
    \hline    
    07 & 304 & 10.41 & \textbf{16.27} & 16.48 \\
    \hline    
    08 & 185 & 2.48 & \textbf{1.66} & 1.44 \\
    \hline
  \end{tabular}
\end{table}


\subsection{Visual Odometry}

\paragraph{Comparing with deep learning based methods.}
\tabref{tab:vo} shows the visual odometry results on KITTI.
For methods that train on monocular videos,
we align the scale of their predicted results with the ground truth by using the 7-DoF optimization.
The results show that the proposed SC-Depth outperforms the previous monocular alternatives,
and it even shows on par performance with the stereo trained method~\cite{li2018undeepvo}.
However, it is not as good as the very recent approach~\cite{zou2020learning} that models the long-term geometry by using the LSTM module.
Besides, the results show that the proposed Pseudo-RGBD SLAM system improves the accuracy significantly over our SC-Depth,
which is contributed to the geometric optimization.
The success of D3VO~\cite{yang2020d3vo} also confirms the importance of geometric optimization for odometry accuracy.
However, note that stereo-trained methods can estimate depths at the metric scale,
which are readily optimized in existing SLAM frameworks.
By contrast, the monocular trained methods suffer from the scale inconsistency issue,
which makes the post-optimization non-trivial---See \figref{fig:keypoints}.
Our contribution here is enabling the monocular trained methods to predict the scale consistent results
so that it allows for optimizing the predicted depths and poses by using the classical geometric frameworks.
A qualitative comparison is provided in \figref{fig:vo},
which shows that the trajectory optimized by our Pseudo-RGBD SLAM is more well-aligned with the ground truth than our SC-Depth and other learning-based methods.


\begin{figure}[t]
\centering
\includegraphics[width=0.95\linewidth]{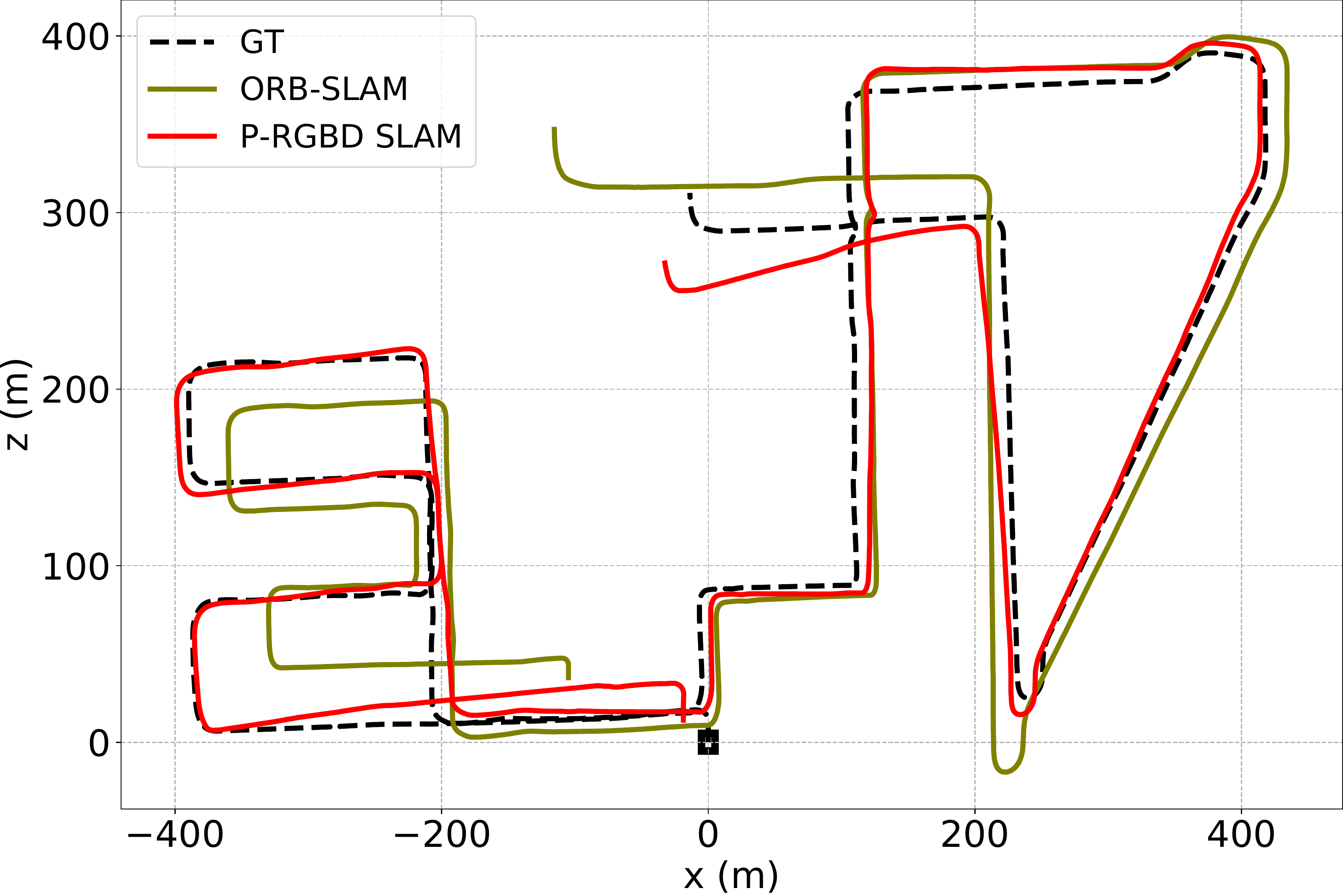} \\
\caption{Estimated trajectory on Seq. 08. 
ORB-SLAM2 is hard to maintain consistent scales over a long video (\eg, left is small, and right is big),
while our method is able by leveraging the scale-consistent depth prediction.
}
\label{fig:scale}
\end{figure}

\begin{figure*}[t]
  \centering
  \addImg[1]{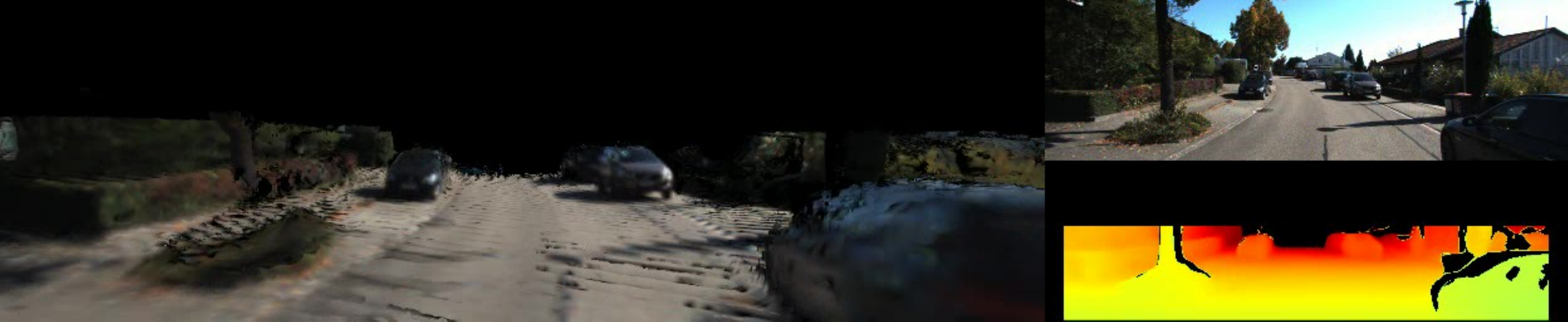}
  \caption{Dense multi-view reconstruction on Seq. 09. 
    The left column shows the reconstructed 3D voxels. 
    The right column shows the input RGB image, estimated depth map.
    We use the depth CNN trained on Seq. 00-08, and the predicted depth is cropped and masked by using our proposed $M_s$. 
  }\label{fig:demo}
\end{figure*}

\paragraph{Depth consistency evaluation.}
We evaluate the geometry consistency of predicted depths by using the point cloud registration metric that is implemented in the Open3D library~\cite{Zhou2018}.
To be specific, we use the ``\emph{open3d.registration.evaluate\_registration}" function.
It computes the RMSE of two aligned point clouds and recognizes the inlier correspondences by a constant threshold.
Then it measures the overlapping area of point clouds by counting the ratio of inlier correspondences in all the target points.
More details can be founded in the Open3D library.
For a given testing sequence, we predict the depth and relative pose for every adjacent image pair,
and we convert the depth into point clouds for evaluation,
where all the depth maps are resized to $832 \times 256$ resolution for a fair comparison.
\tabref{tab:consistency} shows the results,
where we compare our method with Monodepth2~\cite{monodepth2}.
It shows that our predicted depths are significantly more consistent than the latter,
and we hypothesize this is the reason why our method can be readily plugged into the ORB-SLAM2 system while the Monodepth2 fails.
We conduct a more detailed comparison by reporting the number of tracked keypoints in each frame.
The results are shown in \figref{fig:keypoints}.


\paragraph{Pose network or motion model.}
\tabref{tab:vo} shows the results, where using the built-in motion model in ORB-SLAM or using our pose CNN for pose initialization leads to similar performance.
We conjecture the reason is that the motion model is satisfied in most driving scenarios,
where forward motion is dominant.
However, we believe that using the pose network is a more general solution because the constant velocity model is violated when abrupt motion occurs.

\paragraph{Comparing with ORB-SLAM2.}
\tabref{tab:vo_orb} shows the odometry results on KITTI~\cite{Geiger2013IJRR}.
We evaluate results on all frames and on keyframes that are selected by ORB-SLAM2~\cite{murORB2},
since the latter cannot provide results for the full sequence due to unsuccessful initialization or tracking failure.
The results on eleven sequences show that our method either achieves on par accuracy with the ORB-SLAM2 or significantly outperforms the latter.
Besides tracking accuracy, our system is more robust than the ORB-SLAM2.
A detailed comparison is provided in \figref{fig:keypoints},
where our method always tracks more points than the latter (\eg, about $800$ vs $100$).
Moreover, we find that ORB-SLAM2 sometimes suffers from heavy scale drifts in long sequences---See \figref{fig:scale} where ORB-SLAM2 provides inconsistent scales between left and right parts.
In this scenario, our method can maintain a consistent scale over the entire sequence by leveraging the scale-consistent depth prediction.

\paragraph{Using more or less keypoints.} \tabref{tab:vo_orb} shows the ablation study results, where our system with $2K$ keypoints is more accurate than that with $8K$ keypoints.
We conjecture the reason is that using more keypoints would also introduce more outliers,
while the geometric optimization requires only a few accurate sparse points.
We hence recommend users choosing keypoint numbers by considering the trade-off between the system accuracy and robustness (\figref{fig:keypoints}).

\begin{figure}[t]
  \centering
  \includegraphics[width=1\linewidth]{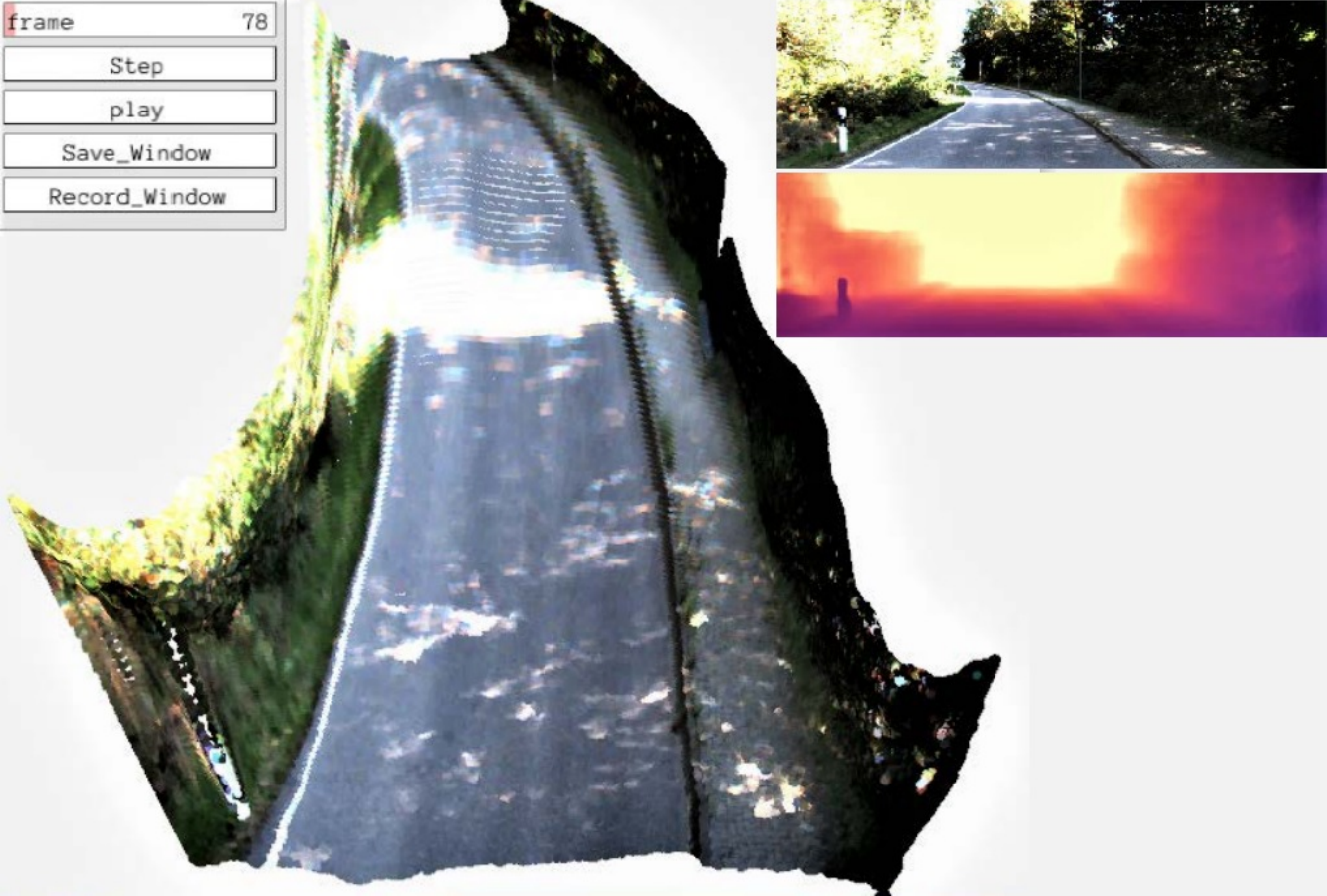}
  \caption{Point cloud visualization on Seq. 09.
  For each incoming image (right 1st row), we predict the depth map (right 2nd row) using our trained network 
  and convert it to a 3D point cloud, 
  which is rendered using the color image and visualized in an eye-bird view (left). 
  }\label{fig:demo-kitti}
  \vspace{-5mm}
\end{figure}

\vspace{-3mm}
\paragraph{Zero-short generalization.}
We validate the generalization ability of our proposed Pseudo-RGBD SLAM on KAIST urban dataset~\cite{jeong2019complex},
where our models are trained on KITTI.
The results are reported in \tabref{tab:kaist}.
It shows that our method consistently outperforms ORB-SLAM2,
which demonstrates the robustness of our proposed system.
Moreover, we demonstrate the generalization ability of our method by presenting a real-world demo---See \figref{fig:demo-adl}.


\subsection{Qualitative Evaluation}\label{sec:demo}

We provide several demos in the supplementary material,
which are briefly described below.

\paragraph{Per-frame 3D visualization.}
\figref{fig:demo-kitti} shows the visualization for predicted depths and textured point clouds on Seq. 09.
We use the model trained on Seq. 00-08.
This demo is to show that the predicted scene or object structure by our trained depth CNN is visually reasonable,
and their scales are consistent over time.
Note that inconsistent prediction would cause flickering videos,
while it is doesn't occur in our demo.

\paragraph{Dense multiple view reconstruction.}
\figref{fig:demo} shows our dense reconstruction demo.
As the depth range is wild in outdoor scenes,
we have to reduce the voxel size of TSDF~\cite{curless1996volumetric} for affording the memory requirement, which degrades the reconstruction quality.
Although the reconstruction is inferior to the state-of-the-art methods,
this demo clearly demonstrates the high consistency of our estimated depths.

\paragraph{Generalization on real-world videos.}
\figref{fig:demo-adl} shows the depth and camera trajectory generated by our method on a self-captured driving video.
The video is captured in Adelaide, Australia.
We use a single camera, which is mounted on a driving car.
Due to the lack of an accurate ground truth trajectory, we use the Google map for qualitative evaluation.
The scene is so challenging that ORB-SLAM2~\cite{murORB2} is unable to generate a complete trajectory,
while the proposed Pseudo-RGBD SLAM performs well.

\section{Conclusion}

This paper proposes a video-based unsupervised depth learning method.
Thanks to the proposed geometry consistency loss and masking scheme,
our trained network can predict scale-consistent and accurate depths over a video.
The depth accuracy is comprehensively evaluated in both indoor and outdoor scenes,
and the quantitative results are attached.
Besides, we demonstrate better consistency against the related work which shows on par depth accuracy to our method,
and we show that such consistency enables our method to be readily plugged into the existing Visual SLAM system.
This shows the possibility of leveraging the depth network that is unsupervised trained from monocular videos for camera tracking and dense reconstruction.

\section*{Acknowledgments}
This work was in part  supported by the Australian Centre of Excellence for Robotic Vision CE140100016, 
and the ARC Laureate Fellowship FL130100102 to Prof. Ian Reid. 
This work was supported by Major Project for New Generation of AI (No. 2018AAA0100403),
Tianjin Natural Science Foundation (No. 18JCYBJC41300 and No. 18ZXZNGX00110), and NSFC (61922046) to Prof. Ming-Ming Cheng.
We also thank anonymous reviewers for their valuable suggestions.

\bibliographystyle{ijcv}
{\footnotesize\bibliography{reference}}

\end{document}